\documentclass[10pt,twocolumn,letterpaper]{article}

\usepackage{cvpr}
\usepackage{times}
\usepackage{epsfig}
\usepackage{graphicx}
\usepackage{amsmath}
\usepackage{amssymb}

\usepackage{caption}
\usepackage{placeins}

\usepackage{xcolor}

\newcommand*\diff{\mathop{}\!\mathrm{d}}
\newcommand\blfootnote[1]{%
  \begingroup
  \renewcommand\thefootnote{}\footnote{#1}%
  \addtocounter{footnote}{-1}%
  \endgroup
}

\usepackage{booktabs}
\usepackage{multirow}
\usepackage{graphbox}

\usepackage[pagebackref=true,breaklinks=true,letterpaper=true,colorlinks,bookmarks=false]{hyperref}

\cvprfinalcopy %

\def\cvprPaperID{1449} %

\ifcvprfinal\fi
\begin{document}
	\parskip = 1pt

	\title{A Variational U-Net for Conditional Appearance and Shape Generation}
	
  \author{Patrick Esser\thanks{Both authors contributed equally to this
  work.}, \space Ekaterina Sutter\footnotemark[1], \space Bj\"orn Ommer\\
		Heidelberg Collaboratory for Image Processing\\
		IWR, Heidelberg University, Germany\\
		{\tt\small firstname.lastname@iwr.uni-heidelberg.de}
	}

	\maketitle
	\thispagestyle{empty}

	\begin{abstract}
    Deep generative models have demonstrated great performance in image
    synthesis. However, results deteriorate in case of spatial deformations,
    since they generate images of objects directly, rather than modeling the
    intricate interplay of their inherent shape and appearance. We present a
    conditional U-Net \cite{unet} for shape-guided image generation,
    conditioned on the output of a variational autoencoder for appearance.
    The approach is trained end-to-end on images, without requiring samples
    of the same object with varying pose or appearance. Experiments show
    that the model enables conditional image generation and transfer.
    Therefore, either shape or appearance can be retained from a query
    image, while freely altering the other. Moreover, appearance can be
    sampled due to its stochastic latent representation, while preserving
    shape. In quantitative and qualitative experiments on COCO \cite{mscoco},
    DeepFashion \cite{deepFashion1,deepFashion2}, shoes~\cite{shoes}, Market-1501
    \cite{market1501} and handbags~\cite{igan}
    the approach demonstrates significant improvements over the state-of-the-art.
	\end{abstract}

    \section{Introduction}	
    
    \begin{figure}
    	\begin{center}
    		\includegraphics[width=0.38\textwidth]{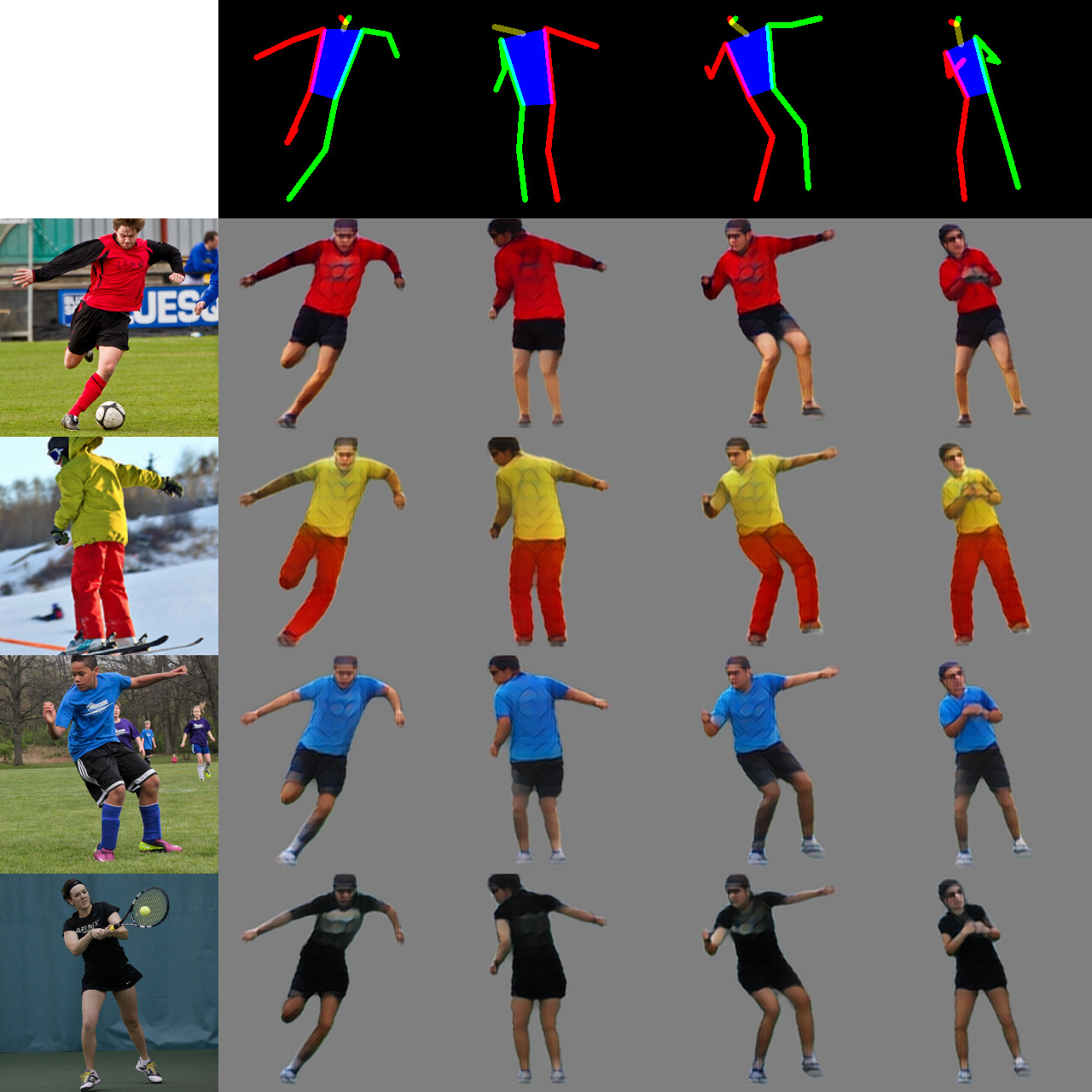}
    	\end{center}
      \caption{\small{Our model learns to infer appearance from the queries
      on the left and can synthesize images with that appearance in
      different poses given in the top row. An animated version can be found
      at \href{https://compvis.github.io/vunet}{https://compvis.github.io/vunet}.}}
    	\label{fig:teaser}
    \end{figure}

    Recently there has been great interest in generative models for image
    synthesis \cite{photographic, pix2pix2016, peopleInClothing,
    PoseGuidedGeneration, igan, cyclegan, rubio:PR:2015}. Generating images of objects
    requires a detailed understanding of both, their appearance and spatial
    layout. Therefore, we have to distinguish basic object characteristics.
    On the one hand, there is the shape and geometrical layout of an object
    relative to the viewpoint of the observer (a person sitting, standing,
    or lying or a folded handbag). On the other hand, there are inherent
    appearance properties such as those characterized by color and texture
    (curly long brown hair vs. buzz cut black hair or the pattern of
    corduroy). Evidently, objects naturally change their shape, while
    retaining their inherent appearance (bending a shoe does not change its
    style). However, the picture of the object varies dramatically in the
    process, e.g., due to translation or even self-occlusion. Conversely,
    the color or fabric of a dress can change with no impact on its shape,
    but again clearly altering the image of the dress.
    
    With deep learning, there has lately been great progress in
    generative models, in particular generative
    adversarial networks (GANs) \cite{wgan, infogan, gan, dcgan,cgan},
    variational autoencoders \cite{vae}, and their combination
    \cite{cvaegan,vaegan}. Despite impressive results, these models still
    suffer from weak performance in case of image distributions with large
    spatial variation: while on perfectly registered faces (e.g., aligned
    CelebA dataset~\cite{celeba}) high-resolution images have been generated
    \cite{srgan, progressivegrowing}, synthesizing the full human body from
    datasets as diverse as COCO \cite{mscoco} is still an open challenge.
    The main reason for this is that these generative models directly
    synthesize the image of an object, but fail to model the intricate
    interplay of appearance and shape that is producing the image.
    Therefore, they can easily add facial hair or glasses to a face as this
    amounts to recoloring of image areas. Contrast this to a person moving
    their arm, which would be represented as coloring the arm at the old
    position with background color and turning the background at the new
    position into an arm. What we are lacking is a generative model that can
    move and deform objects and not only blend their color.
   
   Therefore, we seek to model both, appearance and shape, and their
   interplay when generating images. For general applicability, we want to
   be able to learn from mere still image datasets with no need for a series
   of images of the same object instance showing different articulations. We
   propose a conditional U-Net \cite{unet} architecture for mapping from
   shape to the target image and condition on a latent representation of a
   variational autoencoder for appearance. To disentangle shape and
   appearance, we allow to utilize easily available information related to
   shape, such as edges or automatic estimates of body joint locations. Our
   approach then enables conditional image generation and transfer: to
   synthesize different geometrical layouts or change the appearance of an
   object, either shape or appearance can be retained from a query image,
   whereas the other component can be freely altered or even imputed from
   other images. Moreover, the model also allows to sample from the
   appearance distribution without altering the shape.
\section{Related work}
\label{sec:relatedworks}

In the context of deep learning, three different approaches to image
generation can be identified. Generative Adversarial Networks \cite{gan},
Autoregressive (AR) models \cite{pixelcnndecoder} and Variational
Auto-Encoders (VAE) \cite{vae}.

Our method provides control over both, appearance and shape. In contrast,
many previous methods can control the generative process only with respect
to appearance. \cite{semisup, acgan, cgan} utilize class labels,
\cite{attribute2image} attributes and \cite{stackgan, beYourOwnPrada}
textual descriptions to control the appearance.

Control over shape has been mainly obtained in the Image-to-Image
translation framework. \cite{pix2pix2016} uses a discriminator to obtain
realistic outputs but their method is limited to the synthesis of a single,
uncontrollable appearance. To obtain a larger variety of appearances,
\cite{peopleInClothing} first generates a segmentation mask of fashion
articles and then synthesizes an image. This leads to larger variations in
appearances but does not allow to change the pose of a given appearance.
  
\cite{photographic} uses segmentation masks to produce images in the context
of street scenes as well. They do not rely on adversarial training but
directly learn a multimodal distribution for each segmentation label. The
amount of appearances that can be produced is given by the number of
combinations of modes, resulting in very coarse modeling of appearance. In
contrast, our method makes no assumption that the data can be well
represented by a limited number of modes, does not require segmentation
masks, and it includes an inference mechanism for appearance.
  
\cite{whatandwhere} utilizes the GAN framework and \cite{msar} the
autoregressive framework to provide control over shape and appearance.
However the appearance is specified by very coarse text descriptions.
Furthermore, both methods have problems producing the desired shape
consistently.

In contrast to our generative approach, \cite{cliqueCNN,bautistaCVPR17} have
pursued unsupervised learning of human posture similarity for retrieval in
still images and \cite{milbichICCV17,brattoliCVPR17} in videos.
Rendering images of persons in different poses has been considered by
\cite{personmultiview} for a fixed, discrete set of target poses, and by
\cite{PoseGuidedGeneration} for general poses. In the latter, the authors
use a two-stage model. The first stage implements pixelwise regression to a
target image from a conditional image and the pose of the target image. Thus
the method is fully supervised and requires labeled examples of the same
appearance in different poses. As the result of the first stage is in most cases
too blurry, they use a second stage which employs adversarial training to
produce more realistic images. Our method is never directly trained on the
transfer task and therefore does not require such specific datasets.
Instead, we carefully model the separation between shape and appearance and
as a result, obtain an explicit representation of the appearance which can
be combined with new poses.
\section{Approach}
\label{seq:methodology}

Let $x$ be an image of an object from a dataset $X$. We want to understand
how images are influenced by two essential characteristics of the objects
that they depict: their shape $y$ and appearance $z$. Although the precise
semantics of $y$ can vary, we assume it characterizes geometrical
information, particularly location, shape, and pose. $z$ then represents the
intrinsic appearance characteristics.

If $y$ and $z$ capture all variations of interest, the variance of a
probabilistic model for images conditioned on those two variables is only
due to noise. Hence, the maximum a posteriori estimate $\arg \max_x p(x\vert
y,z)$ serves as an image generator controlled by $y$ and $z$. How can we
model this generator?

\subsection{Variational Autoencoder based on latent shape and appearance}
\label{sec:vae}
If $y$ and $z$ are both latent variables, a popular way of learning the
generator $p(x\vert y,z)$ is to use a VAE. To learn $p(x \vert y,z)$ we need
to maximize the log-likelihood of observed data $x$ and marginalize out the
latent variables $y$ and $z$. To avoid the intractable integral, one
introduces an approximate posterior $q(y, z\vert x)$ to obtain the evidence
lower bound (ELBO) from Jensen's inequality,
\begin{align}
  \log p(x) &= \log \int p(x, y, z) \diff z \diff y \nonumber\\
  &= \log \int \frac{p(x, y, z)}{q(y, z\vert x)} q(y, z\vert x) \nonumber\\
  &\ge \mathbb{E}_{q} \log \frac{p(x\vert y, z)p(y,z)}{q(y, z\vert x)}.
 \label{eq:vae}
\end{align}
As one can see, Eq.~\ref{eq:vae} contains the prior $p(y,z)$, which is
assumed to be a standard normal distribution in the VAE framework. With this
joint prior we cannot guarantee that both variables, $y$ and $z$ would be
separated in the latent space. Thus, our overall goal of separately altering
shape and appearance cannot be met. A standard normal prior can model $z$
but it is not suited to describe the spatial information contained in $y$,
which is localized and easily gets lost in the bottleneck. Therefore, we
need additional information to disentangle y and z when learning the
generator $p(x\vert y, z)$. 

\subsection{Conditional Variational Autoencoder with appearance}
\label{sec:cvae}

In the previous section we have shown that a standard VAE with two latent
variables is not suitable for learning disentangled representations of $y$
and $z$. Instead we assume that we have an estimator function $e$ for the
variable $y$, i.e., $\hat{y}=e(x)$. For example, $e$ could provide
information on shape by extracting edges or automatically estimating body
joint locations \cite{jointestimator,HED}. Following up on
Eq.~\ref{eq:vae}, the task is now to infer the latent variable $z$ from the
image and the estimate $\hat{y}= e(x)$ by maximizing their conditional
log-likelihood.
\begin{align}
	\log p(x\vert \hat{y}) & = \log \int_z p(x, z\vert \hat{y}) \diff z \ge \mathbb{E}_{q} \log \frac{p(x,z\vert \hat{y})}{q(z\vert x, \hat{y})} \nonumber\\
	&= \mathbb{E}_{q} \log \frac{p(x\vert \hat{y},z)p(z\vert \hat{y})}{q(z\vert x,\hat{y})} 
	\label{eq:cvae}
\end{align}

Compared to Eq.~\ref{eq:vae}, the ELBO in Eq.~\ref{eq:cvae} depends now on
the (conditional) prior $p(z\vert\hat{y})$. This distribution can now be
estimated from the training data and captures potential interrelations
between shape and appearance. For instance a person jumping is less likely
to wear a dinner jacket than a T-shirt.

Following~\cite{varappforgan} we model $p(x\vert \hat{y}, z)$ as a
parametric Laplace and $q(z\vert x, \hat{y})$ as a parametric Gaussian
distribution. The parameters of these distributions are estimated by two
neural networks $G_\theta$ and $F_\phi$ respectively. Using the
reparametrization trick~\cite{vae}, these networks can be trained end-to-end
using standard gradient descent. The loss function for training follows
directly from Eq.~\ref{eq:cvae} and has the form:
\begin{align}
\mathcal{L}(x, \theta, \phi) = -KL (&q_\phi(z\vert x, \hat{y}) \vert\vert p_\theta(z\vert\hat{y})) \nonumber \\
 & +\mathbb{E}_{q_\phi(z\vert x,\hat{y})}[\log\ p_\theta(x\vert \hat{y},z)],
\label{eq:vaeloss}
\end{align}
where $KL$ denotes Kullback-Leibler divergence.
The next section derives the network architecture we use for modeling $G_\theta$ and $F_\phi$.

\begin{figure}[t]
	\begin{center}
		\includegraphics[scale=1.0]{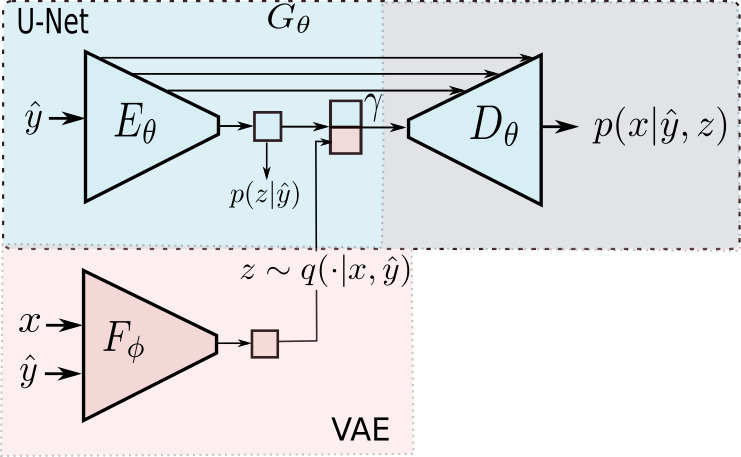}
	\end{center}	
	\caption{\small{Our conditional U-Net combined with a variational autoencoder. $x$: query image, $\hat{y}$: shape estimate, $z$: appearance.}}
	\label{fig:model}
\end{figure}
	
\subsection{Generator}
\label{sec:generator}

	\begin{table*}[t!] \begin{center}
		\begin{tabular}{cc|c|c|ccccc}
			\toprule
			\multicolumn{2}{c|}{GT} & pix2pix\cite{pix2pix2016} & our (reconst.) & \multicolumn{5}{c}{our (random samples)} \\
			\midrule
			\includegraphics[scale=0.13]{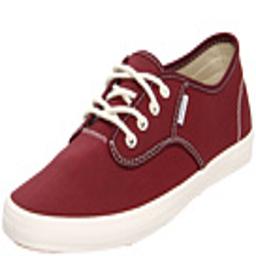} &
			\includegraphics[scale=0.13]{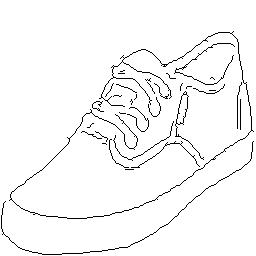} &
			\includegraphics[scale=0.13]{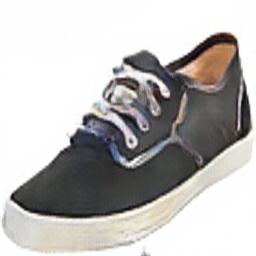} &
			\includegraphics[scale=0.13]{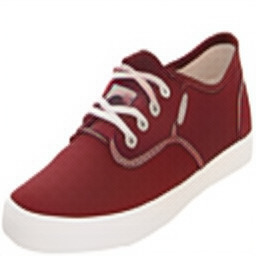} &
			\includegraphics[scale=0.13]{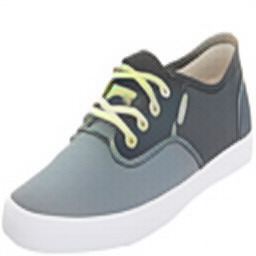} &
			\includegraphics[scale=0.13]{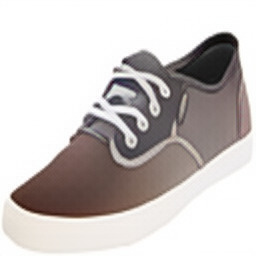} &						\includegraphics[scale=0.13]{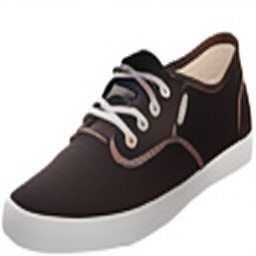} &												\includegraphics[scale=0.13]{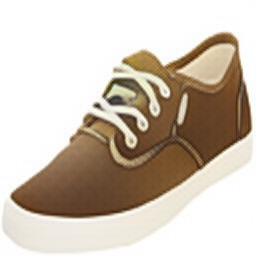} &
			\includegraphics[scale=0.13]{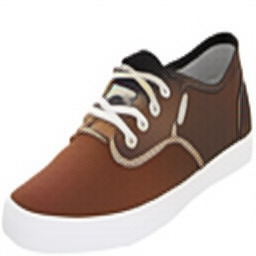} \\			
			\includegraphics[scale=0.13]{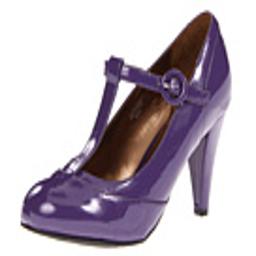} &
			\includegraphics[scale=0.13]{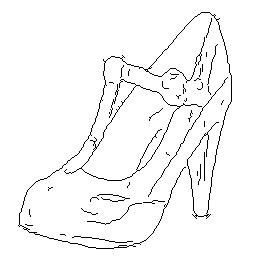} &
			\includegraphics[scale=0.13]{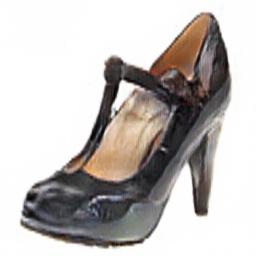} &
			\includegraphics[scale=0.13]{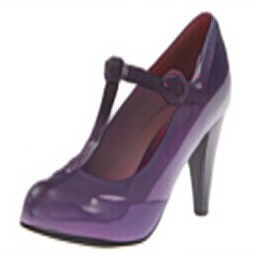} &
			\includegraphics[scale=0.13]{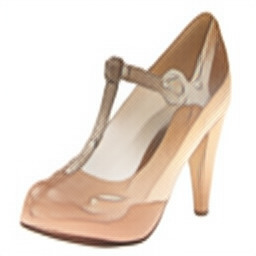} &
			\includegraphics[scale=0.13]{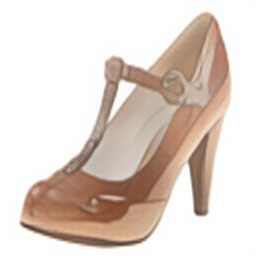} &						\includegraphics[scale=0.13]{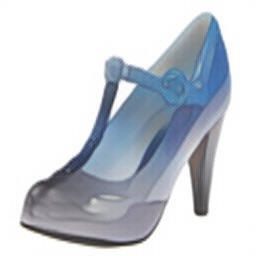} &												\includegraphics[scale=0.13]{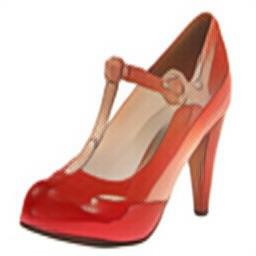} &
			\includegraphics[scale=0.13]{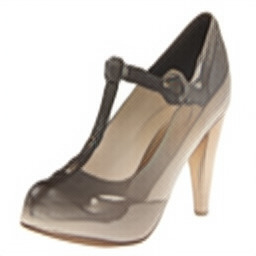} \\				
			\midrule
			\includegraphics[scale=0.13]{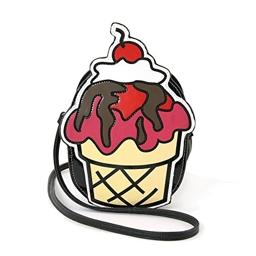} &
			\includegraphics[scale=0.13]{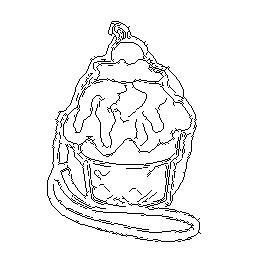} &
			\includegraphics[scale=0.13]{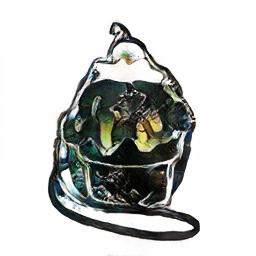} &
			\includegraphics[scale=0.13]{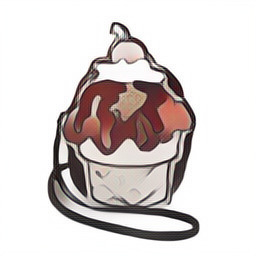} &
			\includegraphics[scale=0.13]{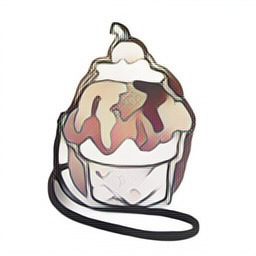} &
			\includegraphics[scale=0.13]{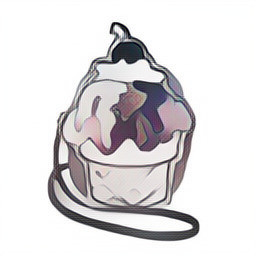} &						\includegraphics[scale=0.13]{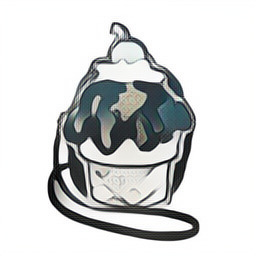} &												\includegraphics[scale=0.13]{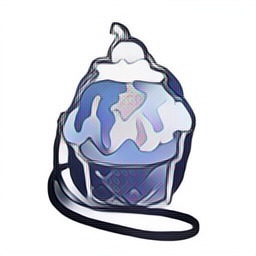} &
			\includegraphics[scale=0.13]{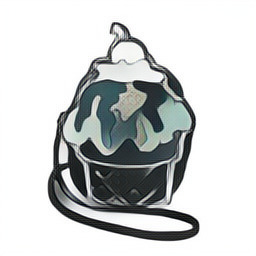} \\			
			\includegraphics[scale=0.13]{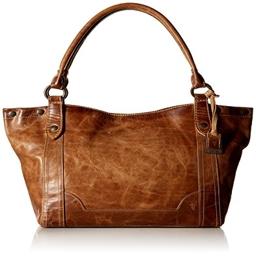} &
			\includegraphics[scale=0.13]{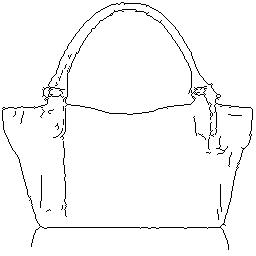} &
			\includegraphics[scale=0.13]{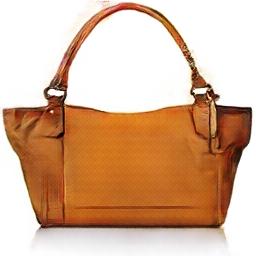} &
			\includegraphics[scale=0.13]{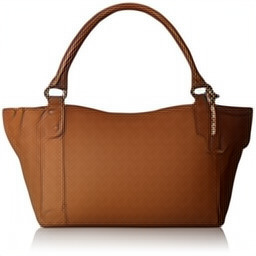} &
			\includegraphics[scale=0.13]{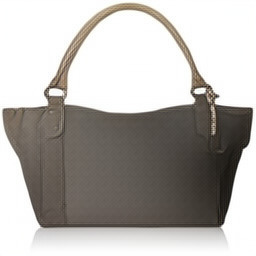} &
			\includegraphics[scale=0.13]{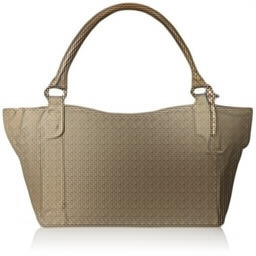} &						\includegraphics[scale=0.13]{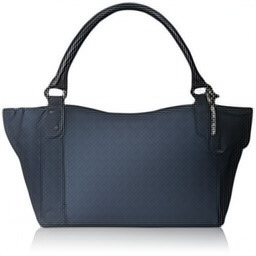} &												\includegraphics[scale=0.13]{images/edges2handbags/test_183_sample_01} &
			\includegraphics[scale=0.13]{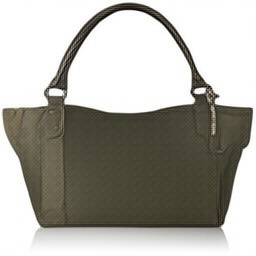} \\				
			\bottomrule		
		\end{tabular}
		\end{center}
    \captionof{figure}{\small{Generating images with only the edge image as
    input (GT image (left) is held back). We compare our approach to pix2pix
    on the datasets of shoes~\cite{shoes} and handbags~\cite{igan}. On the
    right: sampling from our latent appearance distribution.}} 
		\label{fig:edges2images_samples}
	\end{table*}

Let us first establish a network $G_\theta$ which estimates the parameters
of the distribution $p(x\vert \hat{y}, z)$. We assume further, as it is
common practice~\cite{vae}, that the distribution $p(x\vert \hat{y}, z)$ has
constant standard deviation and the function $G_\theta(\hat{y}, z)$ is a
deterministic function in $\hat{y}$. As a consequence, the network
$G_\theta(\hat{y}, z)$ can be considered as an image generator network and
we can replace the second term in Eq.~\ref{eq:vaeloss} with the
reconstruction loss $\mathcal{L}(x, \theta) = \|x - G_\theta(\hat{y},
z)\|_1$:
\begin{align}
\mathcal{L}(x, \theta, \phi) = -KL (&q_\phi(z\vert x, \hat{y}) \vert\vert p_\theta(z\vert\hat{y})) \nonumber \\
 & +\|x - G_\theta(\hat{y}, z)\|_1.
\label{eq:vaeloss_2}
\end{align}
It is well known that pixelwise statistics of images, such as the $L_1$-norm
here, do not model perceptual quality of images well \cite{vaegan}. Instead
we adopt the perceptual loss from \cite{photographic} and formulate the
final loss function as:
\begin{align}
\mathcal{L}(x, \theta, \phi) & = -KL (q_\phi(z\vert x, \hat{y}) \vert\vert p_\theta(z\vert\hat{y})) \nonumber \\
 & +\sum_k\lambda_k\|\Phi_k(x) - \Phi_k(G_\theta(\hat{y}, z))\|_1,
\label{eq:vaeloss_3}
\end{align}
where $\Phi$ is a network for measuring perceptual similarity (in our case
VGG19~\cite{vgg}) and $\lambda_k, k$ are hyper-parameters that control the
contribution of the different layers of $\Phi$ to the total loss.

If we forget for a moment about $z$, the task of the network
$G_\theta(\hat{y})$  is to generate an image $\bar{x}$ given the estimate
$\hat{y}$ of the shape information of an image $x$. Here it is crucial that
we want to preserve spatial information given by $\hat{y}$ in the output
image $\bar{x}$. Therefore, we represent $\hat{y}$ in the form of an image
of the same size as $x$. Depending on the estimate $e:\ e(x)=\hat{y}$ this is
easy to achieve. For example, estimated joints of a human body can be used
to draw a stickman for this person. Given such image representation of
$\hat{y}$ we require that each keypoint of $\hat{y}$ is used to estimate
$\bar{x}$. A U-Net architecture~\cite{unet} would be the most appropriate
choice in this case, as its skip-connections help to propagate the
information directly from input to output. In our case, however, the
generator $G_\theta(\hat{y}, z)$ should learn about images by also
conditioning on $z$.

The appearance $z$ is sampled from the Gaussian distribution $q(z\vert x,
\hat{y})$ whose parameters are estimated by the encoder network $F_\phi$.
Its optimization requires balancing two terms. It has to encode enough
information about $x$ into $z$ such that $p(x \vert \hat{y}, z)$ can
describe the data well as measured by the reconstructions loss in
\eqref{eq:vaeloss_2}. At the same time we penalize a deviation from the
prior $p(z \vert \hat{y})$ by minimizing the Kullback-Leibler divergence
between $q(z\vert x, \hat{y})$ and $p(z \vert \hat{y})$. The design of the
generator $G_\theta$ as a U-Net already guarantees the preservation of
spatial information in the output image. Therefore, any additional
information about the shape encoded in $z$, which is not already contained
in the prior, incurs a cost without providing new information on the
likelihood $p(x \vert \hat{y}, z)$. Thus, an optimal encoder $F_\phi$ must
be invariant to shape. In this case it suffices to include $z$ at the
bottleneck of the generator $G_\theta$.

More formally, let our U-Net-like generator $G_\theta(\hat{y})$ consist of
two parts: an encoder $E_\theta$ and a decoder $D_\theta$ (see
Fig.\ref{fig:model}). We concatenate the inferred appearance representation
$z$ with the bottle-neck representation of $G_\theta$:
$\gamma=[E_\theta(\hat{y}), z]$ and let the decoder $D_\theta(\gamma)$
generate an image from it.
Concatenating the shape and appearance features keeps the
gradients for training the respective encoders $F_\phi$ and $E_\theta$ well separated,
while the decoder $D_\theta$ can learn to combine those representations for an optimal synthesis.
Together $E_\theta$ and
$D_\theta$ build a U-Net like network, which guarantees optimal transfer of
spatial information from input to output images. On the other hand, $F_\phi$
when put together with $D_\theta$ frames a VAE that allows appearance
inference. The prior $p(z\vert \hat{y})$ is estimated by $E_\theta$ just
before it concatenates $z$ into its representation. We train all three
networks jointly by maximizing the loss in Eq.~\ref{eq:vaeloss_3}.

\section{Experiments}
	\label{sec:experiments}

    \begin{table*}[h!]
		\begin{center}
			\begin{tabular}{|l|c|c|c|c|c|c|c|c|}
				\hline
				method & \multicolumn{4}{c|}{Market1501}  & \multicolumn{4}{c|}{DeepFashion}\\
				\hline
				& \multicolumn{2}{c|}{IS}  & \multicolumn{2}{c|}{SSIM} & \multicolumn{2}{c|}{IS}  & \multicolumn{2}{c|}{SSIM}\\
				& mean  & std & mean  & std  & mean & std & mean  & std  \\
				\hline
				real data                   & $3.678$  & $0.274$   & $1.000$ & $0.000$ & $3.415$ & $0.399$  & $1.000$ & $0.000$ \\
				\hline
				PG$^2$ G1-poseMaskedLoss    & $3.326$  &    $-$    & $0.340$ &   $-$  & $2.668$  &  $-$     & $0.779$ &   $-$   \\	
				PG$^2$ G1+D                 & \boldmath{$3.490$}  &    $-$    & $0.283$ &   $-$  & \boldmath{$3.091$}  &  $-$     & $0.761$ &   $-$   \\
				PG$^2$ G1+G2+D              & $3.460$  &    $-$    & $0.253$ &   $-$  & $3.090$  &  $-$     & $0.762$ &   $-$   \\
				\hline
				pix2pix                     &  $2.289$ & $0.0489$ & $0.166$ & $0.060$ & $2.640$  & $0.2171$  & $0.646$ & $0.067$ \\
				\hline
				our                         & $3.214$  & $0.119$   & \boldmath{$0.353$} & $0.097$  & $3.087$  & $0.2394$  & \boldmath{$0.786$} & $0.068$ \\
				\hline
			\end{tabular}
		\end{center}
    \caption{\small{Inception scores (IS) and structured similarities (SSIM)
      of reconstructed test images on DeepFashion and Market1501 datasets.
      Our method outperforms both pix2pix~\cite{pix2pix2016} and
      PG$^2$~\cite{PoseGuidedGeneration} in terms of SSIM. As to IS the
      proposed method performs better than pix2pix and obtains comparable
      results to PG$^2$.}}
		\label{table:visual_quality}
	\end{table*}

We now proof the advantages of the proposed method by showing the results of 
image generation in various datasets with different shape estimators $\hat{y}$.
In addition to visual comparisons with other methods, all results are 
supported by numerical experiments. Code and additional experiments can be found at
\href{https://compvis.github.io/vunet}{https://compvis.github.io/vunet}.
	\begin{table*}[t]
		\centering
		\begin{tabular}{cc|c|c|ccccc}
			\toprule
			\multicolumn{2}{c|}{GT} & pix2pix\cite{pix2pix2016} & our (reconst.) & \multicolumn{5}{c}{our (random samples)} \\
			\midrule
			\includegraphics[scale=0.13]{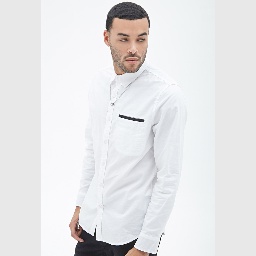} &
			\includegraphics[scale=0.13]{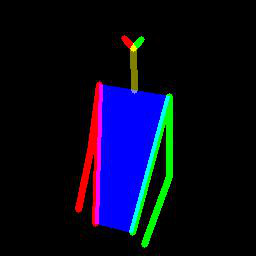} &
			\includegraphics[scale=0.13]{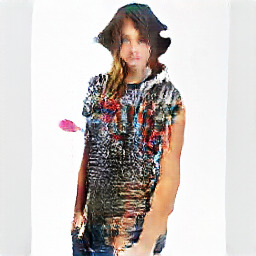} &
			\includegraphics[scale=0.13]{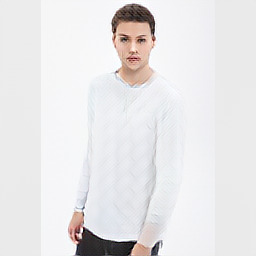} &
			
			\includegraphics[scale=0.13]{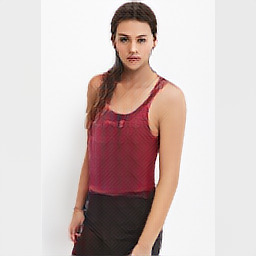} &
			\includegraphics[scale=0.13]{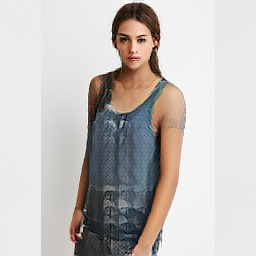} &						\includegraphics[scale=0.13]{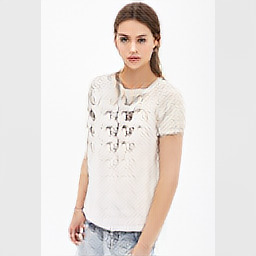} &
			\includegraphics[scale=0.13]{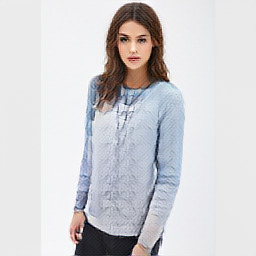} &
			\includegraphics[scale=0.13]{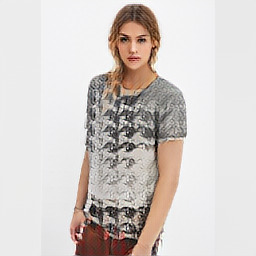} \\	
			\midrule
			\includegraphics[scale=0.3]{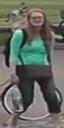} &
			\includegraphics[scale=0.3]{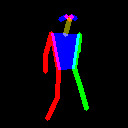} &
			\includegraphics[scale=0.3]{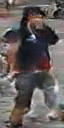} &
			\includegraphics[scale=0.3]{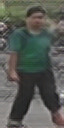} &
			\includegraphics[scale=0.3]{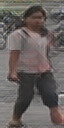} &
			\includegraphics[scale=0.3]{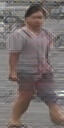} &						\includegraphics[scale=0.3]{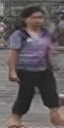} &
			\includegraphics[scale=0.3]{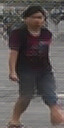} &
			\includegraphics[scale=0.3]{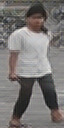} \\			
				
			\bottomrule		
		\end{tabular}
    \captionof{figure}{\small{Generating images based only the stickman as
    input (GT image is held back). We compare our approach with
    pix2pix~\cite{pix2pix2016} on Deepfashion and Market-1501 datasets. On
    the right: sampling from our latent appearance distribution.}} 
		\label{fig:stickman2people_samples}
	\end{table*}

\textbf{Datasets} To compare with other methods, we evaluate on:
shoes~\cite{shoes}, handbags~\cite{igan}, Market-1501~\cite{market1501},
DeepFashion~\cite{deepFashion1,deepFashion2} and COCO~\cite{mscoco}. As
baselines for our subsequent comparisons we use the state-of-the-art pix2pix
model~\cite{pix2pix2016} and PG$^2$~\cite{PoseGuidedGeneration}. To the best
of our knowledge PG$^2$ is the only one approach which is able to transfer
one person to the pose of another. We show that we improve upon this method
and do not require specific datasets for training. With regard to pix2pix,
it is the most general image-to-image translation model which can work with
different shape estimates. Where applicable we directly compare to the
quantitative and qualitative results provided by the authors of the
mentioned papers. As \cite{pix2pix2016} does not perform experiments on
Market-1501, DeepFashion and COCO we train their model on these datasets
using their published code~\cite{pix2pix_page}.

\textbf{Shape estimate} In the following experiments we work with two kinds
of shape estimates: edge images and, in case of humans, automatically
regressed body joint positions. We utilize edges extracted with the HED
algorithm~\cite{HED} by the authors of \cite{pix2pix2016}. Following
\cite{PoseGuidedGeneration} we apply current state-of-the-art real time
multi-person pose estimator~\cite{jointestimator} for body joint regression.

\textbf{Network architecture} The generator $G_\theta$ is implemented as a
U-Net architecture with $2n$ residual blocks~\cite{residual}: $n$ blocks in
the encoder part $E_\theta$ and $n$ symmetric blocks in the decoder part
$D_\theta$. Additional skip-connections link each block in $E_\theta$ to the
corresponding block in $D_\theta$ and guarantee direct information flow from
input to output.
Empirically, we set the parameter $n=7$ which worked well for all considered datasets.
Each residual block follows the
architecture proposed in \cite{residual} without batch normalization. We use
strided convolution with stride $2$ after each residual block to downsample
the input until a bottleneck layer. In the decoder $D_\theta$ we utilize
subpixel convolution~\cite{upsample} to perform the up-sampling between two
consecutive residual blocks. All convolutional layers consists of $3\times3$
filters.
The encoder $F_\phi$ follows the same architecture as the encoder $E_\theta$.

We train our model separately for each dataset using the Adam~\cite{adam}
optimizer with parameters $\beta_1=0.5$ and $\beta_2=0.9$ for $100K$
iterations. The initial learning rate is set to $0.001$ and linearly
decreases to $0$ during training. We utilize weight normalization and data
dependent initialization of weights as described in \cite{weightnorm}. Each
$\lambda_k$ is set to the reciprocal of the total number of elements in
layer $k$.

\textbf{In-plane normalization} In some difficult cases, e.g. for
datasets with high shape variability, it is difficult to perform appearance
transfer from one object to another with no part correspondences between
them. This problem is especially problematic when generating human beings.
To cope with it we propose to use additional in-plane normalization
utilizing the information provided by the shape estimate $\hat{y}$. In our
case $\hat{y}$ is given by the positions of body joints which we
use to crop out areas around body limbs.
This results in $8$ image crops that we stack together and give as
input to the generator $F_\phi$ instead of $x$. If some limbs are missing
(e.g. due to occlusions) we use a black image instead of the corresponding
crop.

	\begin{table}[t]
		\centering
		\begin{tabular}{ccc|ccc}
			\bfseries Input & \bfseries pix2pix & \bfseries Our & \bfseries Input & \bfseries pix2pix & \bfseries Our \\
			\toprule

			\includegraphics[scale=0.12]{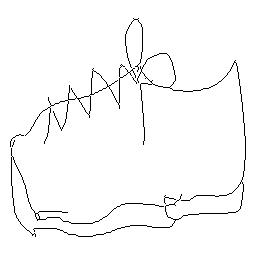} &
			\includegraphics[scale=0.12]{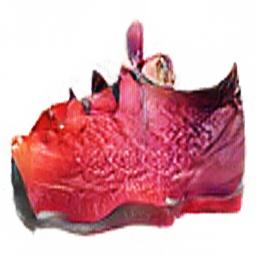} &
			\includegraphics[scale=0.12]{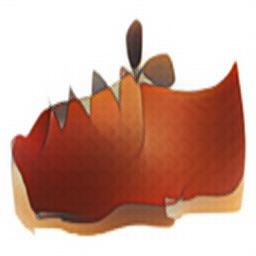} &
			
			\includegraphics[scale=0.12]{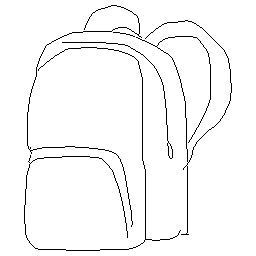} & 
			\includegraphics[scale=0.12]{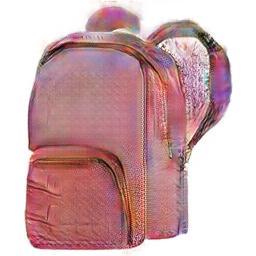} &
			\includegraphics[scale=0.12]{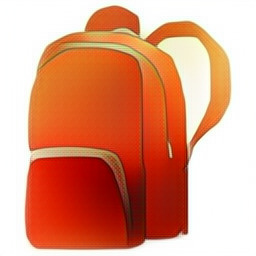} \\
			
			\includegraphics[scale=0.12]{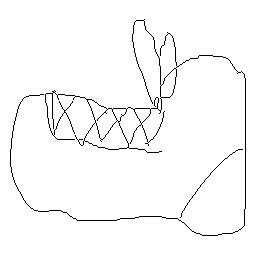} & 
			\includegraphics[scale=0.12]{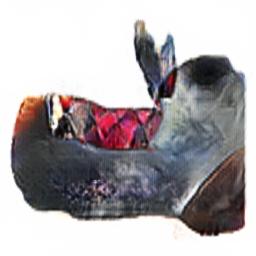} &
			\includegraphics[scale=0.12]{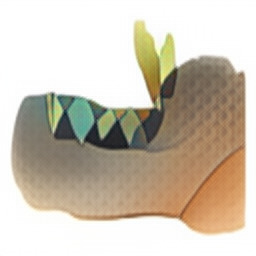} &
			
			\includegraphics[scale=0.12]{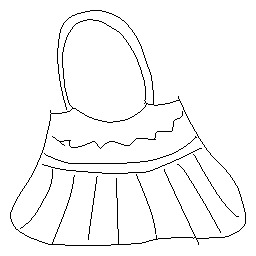} & 
			\includegraphics[scale=0.12]{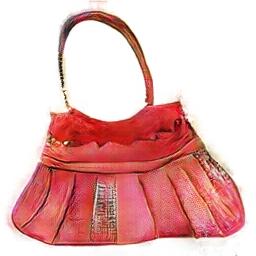} &
			\includegraphics[scale=0.12]{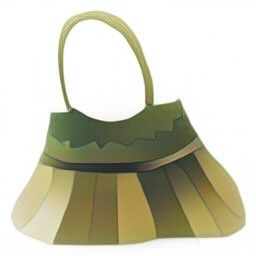} \\
			
			\includegraphics[scale=0.12]{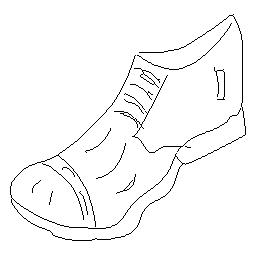} & 
			\includegraphics[scale=0.12]{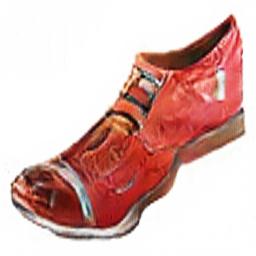} &
			\includegraphics[scale=0.12]{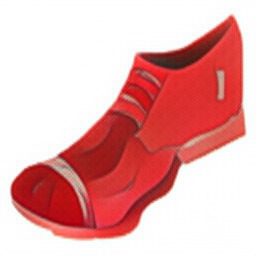} &
			
			\includegraphics[scale=0.12]{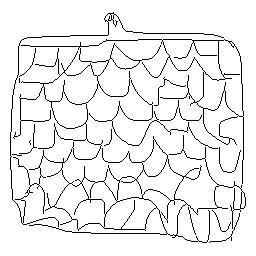} & 
			\includegraphics[scale=0.12]{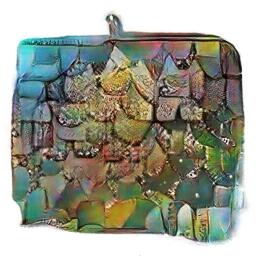} &
			\includegraphics[scale=0.12]{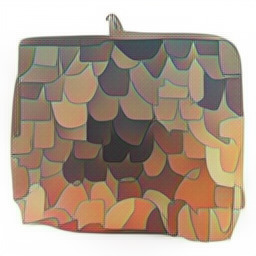} \\
			\bottomrule
			
		\end{tabular}
    \captionof{figure}{\small{Colorization of sketches: we compare
    generalization ability of pix2pix~\cite{pix2pix2016} and our model
    trained on real images. The task is to generate plausible appearances
    for human-drawn sketches of shoes and handbags~\cite{sketches}.}} 
		\label{fig:sketches2images}
	\end{table}
	
Let us now investigate the proposed model for conditional image generation
based on three tasks: 1) reconstruction of an image $x$ given its shape
estimate $\hat{y}$ and original appearance $z$; 2) conditional image
generation based on a given shape estimate $\hat{y}$; 3) conditional image
generation from arbitrary combinations of $\hat{y}$ and $z$. 

	\subsection{Image reconstruction}
	
	\label{seq:imreconstruction}
  Given a query image $x$ and its shape estimate $\hat{y}$ we can use the
  network $F_\phi$ to infer appearance of the image $x$. Namely, we denote
  the mean of the distribution $q(z\vert x, \hat{y})$ predicted by $F_\phi$
  from the single image $x$ as its original appearance $z$. Using these $z$
  and $\hat{y}$ we can ask our generator $G_\theta$ to reconstruct $x$ from
  its two components. 
	
  We show examples of images reconstructed by our methods in
  Figs.~\ref{fig:edges2images_samples} and
  \ref{fig:stickman2people_samples}. Additionally, we follow the experiment
  in~\cite{PoseGuidedGeneration} and calculate for the reconstructions of
  the test images in Market-1501 and DeepFashion dataset Structural
  Similarities (SSIM)~\cite{ssim} and Inception Scores (IS)~\cite{inception}
  (see Table~\ref{table:visual_quality}). Compared to
  pix2pix~\cite{pix2pix2016} and PG$^2$~\cite{PoseGuidedGeneration} our
  method outperforms both in terms of SSIM score. Note that SSIM compares
  the reconstructions directly against the original images. As our method
  differs from both by generating images conditioned on shape and appearance
  this underlines the benefit of this conditional representation for image
  generation. In contrast to SSIM, inception score is measured on the set of
  reconstructed images independently from the original images. In terms of
  IS we achieve comparable results to~\cite{PoseGuidedGeneration} and
  improve on~\cite{pix2pix2016}.

	\subsection{Appearance sampling}
	\label{seq:imsampling}
  An important advantage of our model compared to \cite{pix2pix2016} and
  \cite{PoseGuidedGeneration} is its ability to generate multiple new images
  conditioned only on the estimate of an object's shape $\hat{y}$. This is
  achieved by randomly sampling $z$ from the learned prior $p(z\vert
  \hat{y})$ instead of inferring it directly from an image $x$. Thus,
  appearance can be explored while keeping shape fixed.

  \textbf{Edges-to-images} We compare our method to pix2pix by generating
  images from edge images of shoes or handbags. The results can been seen in
  Fig.~\ref{fig:edges2images_samples}. As noted by the authors in
  \cite{pix2pix2016}, the outputs of pix2pix show only marginal diversity at
  test time, thus looking almost identical. To save space, we therefore
  present only one of them. In contrast, our model generates high-quality
  images with large diversity. We also observe that our model generalizes
  better to sketchy drawings made by humans~\cite{sketches} (see
  Fig.~\ref{fig:sketches2images}). Due to a higher abstraction level,
  sketches are quite different to the edges extracted from the real images
  in the previous experiment. In this challenging task our model shows
  higher coherence to the input edge image as well as less artifacts such as
  at the carrying strap of the backpack.
			
  \textbf{Stickman-to-person} Here we evaluate our model on the task of
  learning plausible appearances for rendering human beings. Given a
  $\hat{y}$ we thus sample $z$ and infer $x$. We compare our results with
  the ones achieved by pix2pix on Market-1501 and DeepFashion datasets (see
  Fig.~\ref{fig:stickman2people_samples}). Due to marginal diversity in the
  output of pix2pix we again only show one sample per row. We observe that
  our model has learned a significantly more natural latent representation
  of the distribution of appearance. Also it preserves the spatial
  layout of the human figure better. We prove this observation by re-estimating
  joint positions from the test images generated by each methods on all
  three datasets. For this we apply the same the algorithm we used to
  estimate the positions of body joints initially,
  namely~\cite{jointestimator} with parameter kept fixed. We report mean
  $L_2$-error in the positions of detected joints in
  Table~\ref{table:pose_preserving}. Our approach shows a significantly
  lower re-localization error, thus demonstrating that body pose has been
  favorably retained.
	
	\begin{figure}
        \begin{center}
    		\includegraphics[scale=0.3]{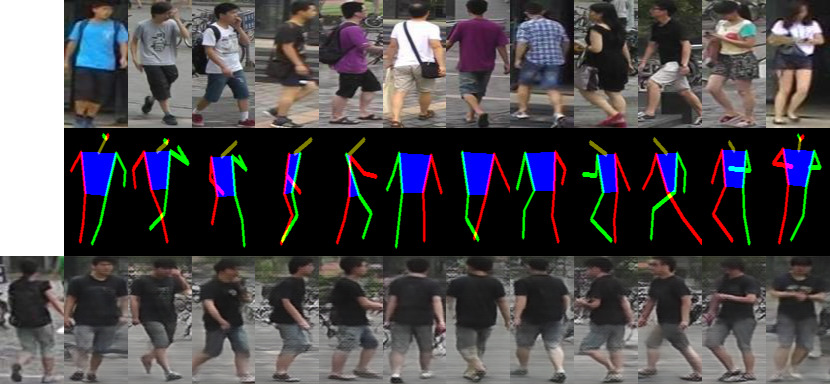}
    	\end{center}
    \caption{\small{Appearance transfer on Market-1501. Appearance is
    provided by image on bottom left. $\hat{y}$ (middle) is automatically
    extracted from image at the top and transferred to bottom.}}
    \label{fig:transfer_market1501}
    \end{figure}
	
	\begin{table}[h!]
		\begin{center}
			\begin{tabular}{|l|ccc|}
				\hline
				\small{method} & \small{our} & \small{pix2pix}  & \small{PG$^2$} \\
				\hline
				\small{COCO}     & \boldmath{$23.23$} & $59.26$ & $-$\\
				\small{DeepFashion} & \boldmath{$7.34$} & $15.53$ & $19.04$ \\ %
				\small{Market1501}  & \boldmath{$54.60$} & $59.59$ & $59.95$ \\ %
				\hline
			\end{tabular}
		\end{center}
    \caption{\small{Automatic body joint detection is applied to images of
    humans synthesized by our method, pix2pix, and PG$^2$. The L2 error of
    joint location is presented, indicating how good shape is preserved.
    The error is measured in pixels based on a resolution of
    $256\times 256$.}}
		\label{table:pose_preserving}
	\end{table}	

    \begin{figure}
    	\begin{center}
    		\includegraphics[width=0.4\textwidth]{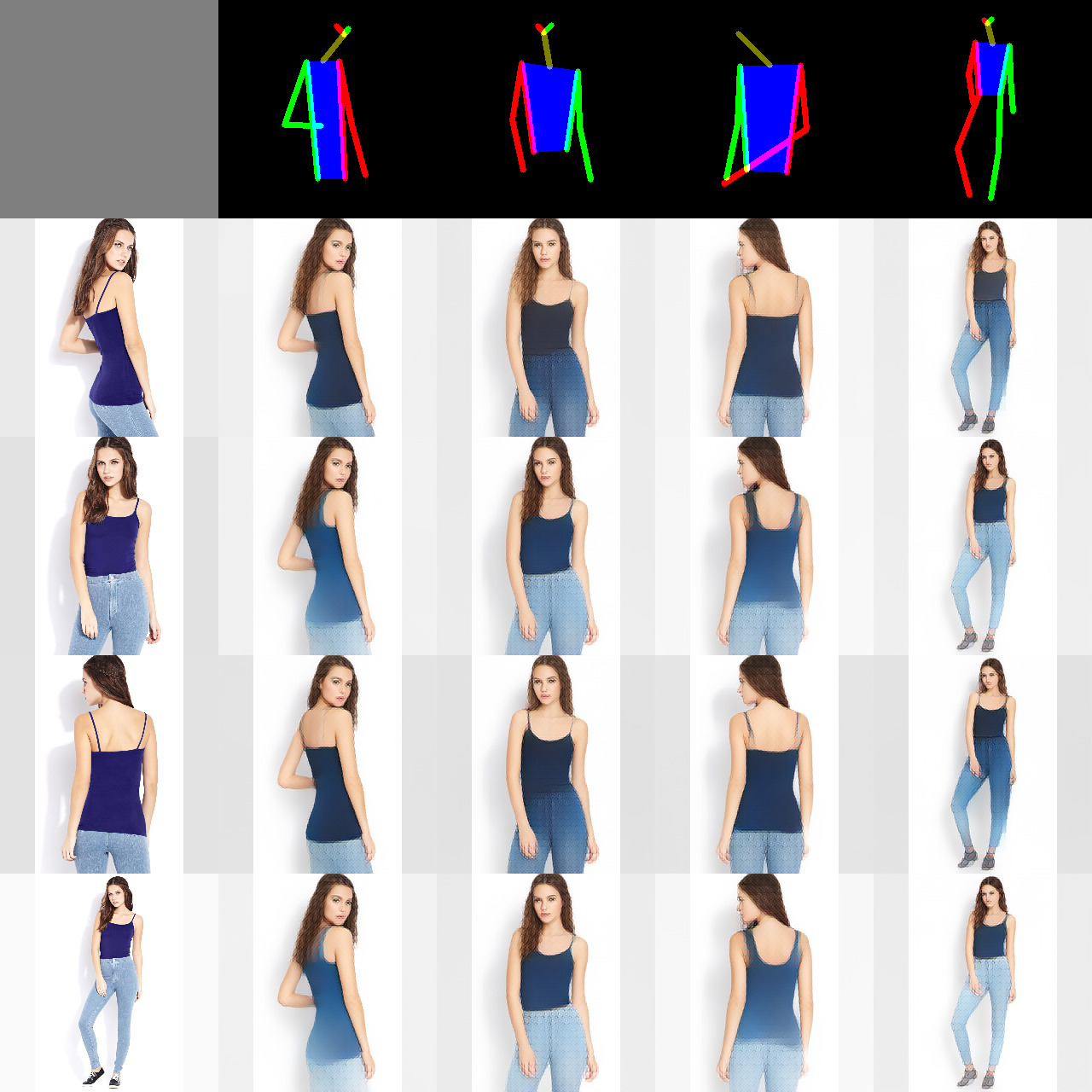}
    	\end{center}
    	\caption{\small{Stability of appearance transfer on DeepFashion. Each row is
      synthesized using appearance information from the leftmost image and
      each column is synthesized from the pose in the first row. Notice that
      inferred appearance remains constant across a wide variety of
      viewpoints.}}
    	\label{fig:stability}
    \end{figure}
    
\subsection{Independent transfer of shape and appearance}
\label{seq:transfer}
We show performance of our method for conditional image transfer,
Fig.~\ref{fig:stability}. Our disentangled representation of shape and
appearance can transfer a single appearance over different shapes and vice
versa. The model has learned a disentangled representation of both
characteristics, so that one can be freely altered without affecting the
other. This ability is further demonstrated in
Fig.~\ref{fig:transfer_market1501} that shows a synthesis across a full
$360^\circ$ turn. 

	\begin{table}
		\begin{center}
			\begin{tabular}{|l|c|c|c|c|}
				\hline
				\small{dataset} & \multicolumn{2}{c|}{\small{Our}} & \multicolumn{2}{c|}{\small{PG$^2$}} \\
				           & $\|std\|$ & \small{max pairwise} & $\|std\|$ & \small{max pairwise} \\
				                  &   & \small{dist} &  & \small{dist} \\
				\hline
				\small{market1501}   & \boldmath{$55.95$}  & \boldmath{$125.99$} & $67.39$  & $155.16$\\
				\small{deepfashion}  & \boldmath{$59.24$}  & \boldmath{$135.83$} & $69.57$  & $149.66$\\
				\small{deepfashion}  & \boldmath{$56.24$}  & \boldmath{$121.47$} & $59.73$    & $127.53$\\
				\hline
			\end{tabular}
		\end{center}
    \caption{\small{Given an image its appearance is transferred from an
    image to different target poses. For these synthesized images, the
    unwanted deviation in appearance is measured using a pairwise perceptual
    VGG16 loss.}}
		\label{table:app_preserving}
	\end{table}

      \begin{table*}
    \centering

    \begin{tabular}{cccccccc}
      \multicolumn{4}{c}{Market} & \multicolumn{4}{c}{DeepFashion} \\
      \cmidrule(lr){1-4}\cmidrule(lr){5-8}
      \bfseries \small{Conditional} & \bfseries \small{Target} & \bfseries \small{Stage} & \bfseries \small{Our} &
      \bfseries \small{Conditional} & \bfseries \small{Target} & \bfseries \small{Stage} & \bfseries \small{Our} \\

      \bfseries \small{image}       & \bfseries \small{image}  & \bfseries \small{II\cite{PoseGuidedGeneration}}&                &
      \bfseries \small{image}       & \bfseries \small{image}  & \bfseries \small{II\cite{PoseGuidedGeneration}}& \\	

      \cmidrule(lr){1-4}\cmidrule(lr){5-8}
      \includegraphics[scale=0.39]{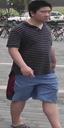} &
      \includegraphics[scale=0.39]{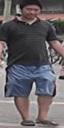} &
      \includegraphics[scale=0.39]{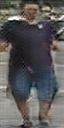} &
      \includegraphics[scale=0.39]{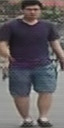} &
      \includegraphics[width=0.10\textwidth,height=0.10\textwidth]{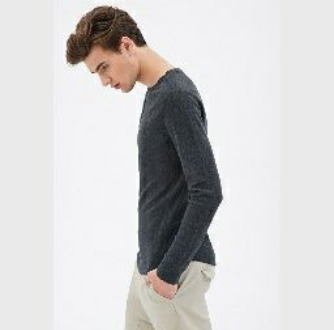} &
      \includegraphics[width=0.10\textwidth,height=0.10\textwidth]{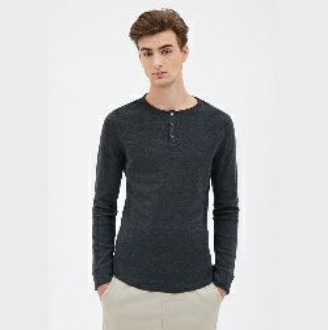} &
      \includegraphics[width=0.10\textwidth,height=0.10\textwidth]{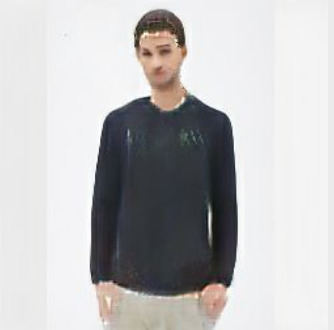} &
      \includegraphics[width=0.10\textwidth,height=0.10\textwidth]{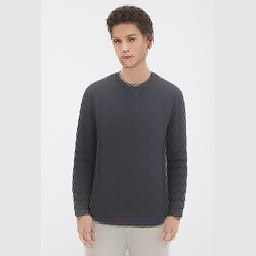} \\

      \includegraphics[scale=0.39]{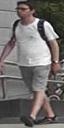} &
      \includegraphics[scale=0.39]{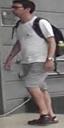} &
      \includegraphics[scale=0.39]{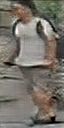} &
      \includegraphics[scale=0.39]{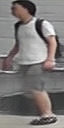} &
      \includegraphics[width=0.10\textwidth,height=0.10\textwidth]{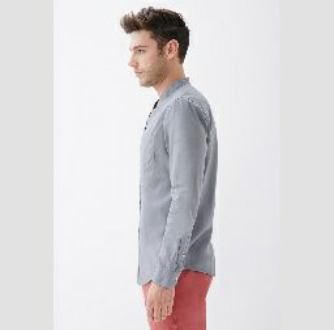} &
      \includegraphics[width=0.10\textwidth,height=0.10\textwidth]{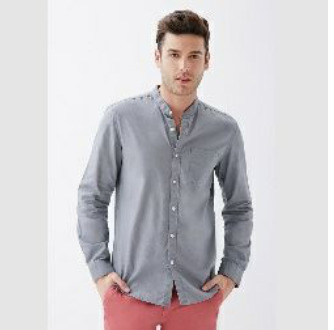} &
      \includegraphics[width=0.10\textwidth,height=0.10\textwidth]{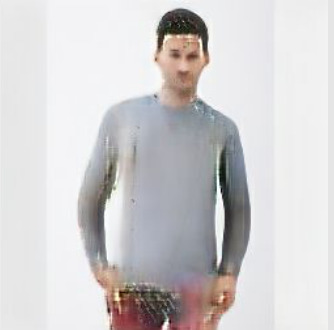} &
      \includegraphics[width=0.10\textwidth,height=0.10\textwidth]{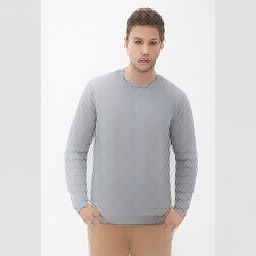} \\

      \includegraphics[scale=0.39]{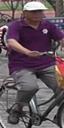} &
      \includegraphics[scale=0.39]{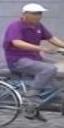} &
      \includegraphics[scale=0.39]{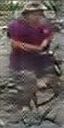} &
      \includegraphics[scale=0.39]{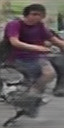} &
      \includegraphics[width=0.10\textwidth,height=0.10\textwidth]{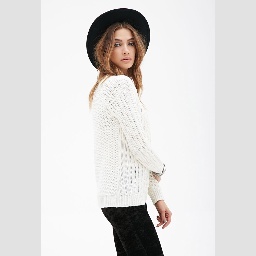} &
      \includegraphics[width=0.10\textwidth,height=0.10\textwidth]{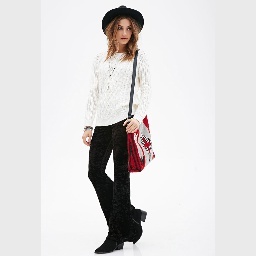} &
      \includegraphics[width=0.10\textwidth,height=0.10\textwidth]{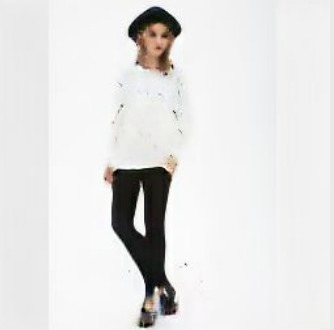} &
      \includegraphics[width=0.10\textwidth,height=0.10\textwidth]{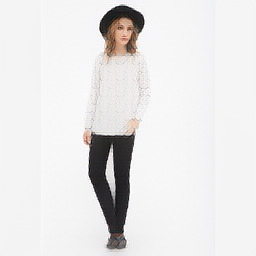} \\

      \includegraphics[scale=0.39]{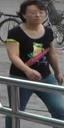} &
      \includegraphics[scale=0.39]{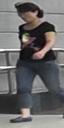} &
      \includegraphics[scale=0.39]{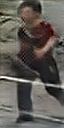} &
      \includegraphics[scale=0.39]{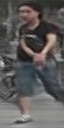} &
      \includegraphics[width=0.10\textwidth,height=0.10\textwidth]{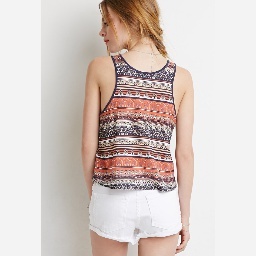} &
      \includegraphics[width=0.10\textwidth,height=0.10\textwidth]{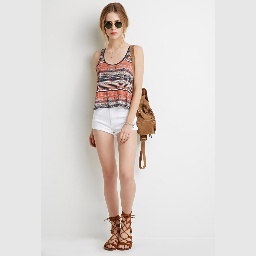} &
      \includegraphics[width=0.10\textwidth,height=0.10\textwidth]{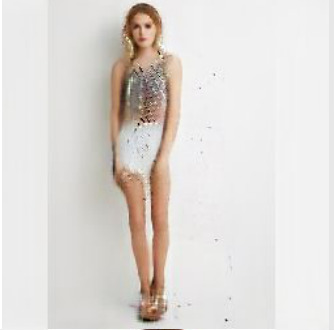} &
      \includegraphics[width=0.10\textwidth,height=0.10\textwidth]{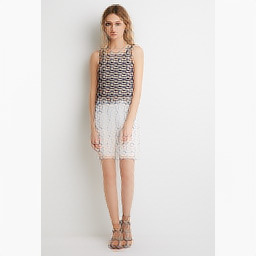} \\
    \end{tabular}
    \captionof{figure}{\small{Comparing image transfer against PG$^2$. Left:
        Results on Market. Right: Results on DeepFashion. Appearance is
        inferred from the conditional image, the pose is inferred from the
        target image. Note that our method does not require labels about
        person identity.}}
    \label{fig:transfer_pg2}
\end{table*}

The only other work we can compare with in this experiment is PG$^2$ from
\cite{PoseGuidedGeneration}. In contrast to our method PG$^2$ was trained
fully supervised on DeepFashion and Market-1501 datasets with pairs of
images that share appearance (person id) but contain different shapes (in
this case pose) of the same person. Despite the fact that we never train our
model explicitly on pairs of images, we demonstrate both qualitatively and
quantitatively that our method improves upon \cite{PoseGuidedGeneration}. A
direct visual comparison is shown in Fig.~\ref{fig:transfer_pg2}. We further
design a new metric to evaluate and compare against PG$^2$ on the appearance
and shape transfer. Since code for \cite{PoseGuidedGeneration} is not
available our comparison is limited to generated images provided by
~\cite{PoseGuidedGeneration}. The idea behind our metric is to compare how
good an appearance $z$ of a reference image $x$ is preserved when
synthesizing it with a new shape estimate $\hat{y}$. For that we first
fine-tune an ImageNet~\cite{ILSVRC15} pretrained VGG16~\cite{vgg} on
Market-1501 on the challenging task of person re-identification. In test
phase this network achieves mean average precision (mAP) of $35.62\%$ and
rank-1 accuracy of $63.00\%$ on a task of single query retrieval. These
results are comparable to those reported in
\cite{zheng2016discriminatively}. Due to the nature of Market-1501, which
contains images of the same persons from multiple viewpoints, the features
learned by the network should be pose invariant and mostly sensitive to
appearance. Therefore, we use a difference between two features extracted by
this network as a measure for appearance similarity.

For all results on DeepFashion and Market-1501 datasets reported in
\cite{PoseGuidedGeneration} we use our method to generate exactly the same
images. Further we build groups of images sharing the same appearance and
retain those groups that contain more than one element. As a result we
obtain three groups of images (see Table.~\ref{table:app_preserving}) which
we analyze independently. We denote these groups with $I_i, i=\{1,2,3\}$. 

For each image $j$ in the group $I_i$ we find its $10$ nearest neighbors
$n^i_{j_1}, n^i_{j_2}, \dots n^i_{j_{10}}$ in the training set using the
embedding of the fine-tuned VGG16. We search for the nearest neighbors in the
training dataset, as the person IDs and poses were taken from the test
dataset. We calculate the mean over each nearest-neighbor set and use this
mean $m_j$ as the unique representation of the generated image $j$. For
images $j$ in the group $I_i$ we calculate maximal pairwise distance between
the $m_j$ as well as the length of the standard deviation vector. The
results over all three image groups $I_1, I_2, I_3$ are summarized in
Table~\ref{table:app_preserving}. One can see that our method shows higher
compactness of the feature representations $m_j$ of the images in each
group. From these results we conclude that our generated images are more
consistent in their appearance than the results of PG$^2$.

\textbf{Generalization to different poses}
Because we are not limited by the availability of labeled images
showing the same appearance in different poses, we can utilize additional
large scale datasets. Results on COCO are shown in Fig.~\ref{fig:teaser}.
Besides still images, we are able to synthesize videos.
Examples can be found at
\href{https://compvis.github.io/vunet}{https://compvis.github.io/vunet},
demonstrating the
transfer of appearances from COCO to poses obtained from a video dataset
\cite{pennaction}.

\subsection{Ablation study}
\label{seq:Ablation}
At last we analyze the effect of individual components of our method on the
quality of generated images (see Fig.~\ref{fig:klnokl}). 

\textbf{Absence of appearance} Without appearance information $z$ our
generator $G_\theta$ is a U-Net performing a direct mapping from shape
estimate $\hat{y}$ to the image $x$. In this case, the output of the
generator is the mean of $p(x \vert y)$. Because we model it as a unimodal
Laplace distribution, it is an estimate of the mean image over all possible
images (of the dataset) with the given shape. As a result the output
generations do not show any appearance at all (Fig.~\ref{fig:klnokl}, second
row).

\textbf{Importance of KL-loss} We show further what happens if we replace
the VAE in our model with a simple autoencoder. In practice that means that
we ignore the KL-term in the loss function in Eq.~\ref{eq:vaeloss_3}. In
this case, the network has no incentive to learn a shape invariant
representation of the appearance and just learns to copy and paste the
appearance inputs to the positions provided by the shape estimate $\hat{y}$
(Fig.~\ref{fig:klnokl}, third row).

\textbf{Our full model} The last row in Fig.~\ref{fig:klnokl} shows that our full model can successfully perform appearance transfer.

\begin{table}
  \centering
  \begin{tabular}{cc|ccc}
        \vtop{\hbox{\strut \small{KL}}} &
    \vtop{\hbox{\strut \small{Appearance}}\hbox{\strut \small{Input}}} &
    \includegraphics[align=c,width=0.09\textwidth,height=0.09\textwidth]{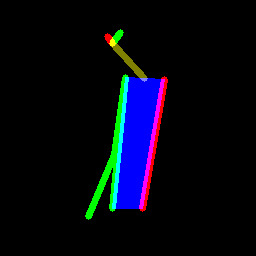} &
    \includegraphics[align=c,width=0.09\textwidth,height=0.09\textwidth]{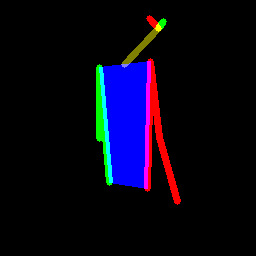} &
    \includegraphics[align=c,width=0.09\textwidth,height=0.09\textwidth]{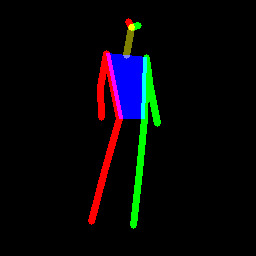} \\
    \midrule
    no & no &
    \includegraphics[align=c,width=0.09\textwidth,height=0.09\textwidth]{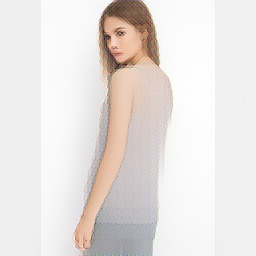} &
    \includegraphics[align=c,width=0.09\textwidth,height=0.09\textwidth]{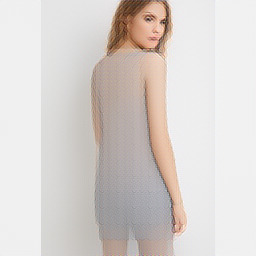} &
    \includegraphics[align=c,width=0.09\textwidth,height=0.09\textwidth]{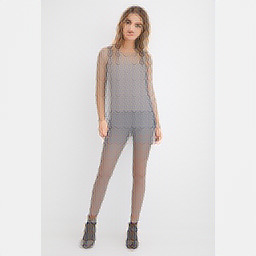} \\
    \midrule
    no &
    \includegraphics[align=c,width=0.09\textwidth,height=0.09\textwidth]{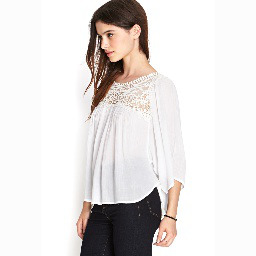} &
    \includegraphics[align=c,width=0.09\textwidth,height=0.09\textwidth]{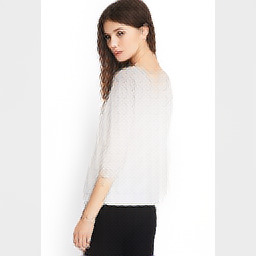} &
    \includegraphics[align=c,width=0.09\textwidth,height=0.09\textwidth]{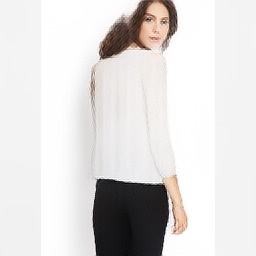} &
    \includegraphics[align=c,width=0.09\textwidth,height=0.09\textwidth]{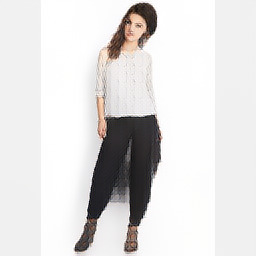} \\
    \midrule
    yes &
    \includegraphics[align=c,width=0.09\textwidth,height=0.09\textwidth]{images/deepfashion/ablation/nobox_style} &
    \includegraphics[align=c,width=0.09\textwidth,height=0.09\textwidth]{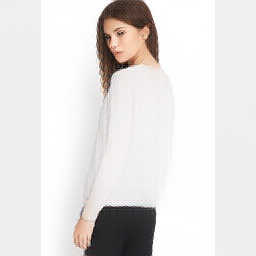} &
    \includegraphics[align=c,width=0.09\textwidth,height=0.09\textwidth]{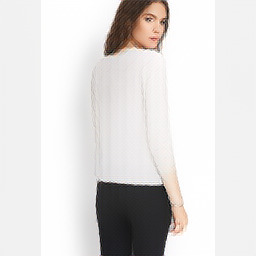} &
    \includegraphics[align=c,width=0.09\textwidth,height=0.09\textwidth]{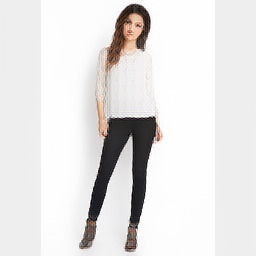} \\
  \end{tabular}
\captionof{figure}{\small{Ablation study on the task of appearance transfer.
  See Sec.~\ref{seq:Ablation}.
  }}
    \label{fig:klnokl}
\end{table}

\section{Conclusion}
We have presented a variational U-Net for conditional image generation by
modeling the interplay of shape and appearance. While a variational
autoencoder allows to sample appearance, the U-Net preserves object shape.
Experiments on several datasets and diverse objects have demonstrated that
the model significantly improves the state-of-the-art in conditional image
generation and transfer.
\blfootnote{This work has been supported in part by the Heidelberg Academy of Science and a
hardware donation from NVIDIA.}

\FloatBarrier
\newpage

{\small
	\bibliographystyle{ieee}
	\bibliography{egbib}

\begin{thebibliography}{10}\itemsep=-1pt

\bibitem{wgan}
M.~Arjovsky, S.~Chintala, and L.~Bottou.
\newblock Wasserstein {GAN}.
\newblock {\em arXiv preprint arXiv:1701.07875}, 2016.

\bibitem{cvaegan}
J.~Bao, D.~Chen, F.~Wen, H.~Li, and G.~Hua.
\newblock {CVAE{-}GAN}: Fine{-}grained image generation through assymetric
  training.
\newblock In {\em To appear in Proceedings of the International Conference on
  Computer Vision (ICCV)}, 2017.

\bibitem{bautistaCVPR17}
M.~Bautista, A.~Sanakoyeu, and B.~Ommer.
\newblock Deep unsupervised similarity learning using partially ordered sets.
\newblock In {\em The IEEE Conference on Computer Vision and Pattern
  Recognition (CVPR)}, 2017.

\bibitem{cliqueCNN}
M.~Bautista, A.~Sanakoyeu, E.~Sutter, and B.~Ommer.
\newblock Cliquecnn: Deep unsupervised exemplar learning.
\newblock In {\em Proceedings of the Conference on Advances in Neural
  Information Processing Systems (NIPS)}, Barcelona, 2016. MIT Press, MIT
  Press.

\bibitem{brattoliCVPR17}
B.~Brattoli, U.~B{\"u}chler, A.~S. Wahl, M.~E. Schwab, and B.~Ommer.
\newblock Lstm self-supervision for detailed behavior analysis.
\newblock In {\em Proceedings of the IEEE Conference on Computer Vision and
  Pattern Recognition (CVPR)}. (BB and UB contributed equally), (BB and UB
  contributed equally), 2017.

\bibitem{jointestimator}
Z.~Cao, T.~Simon, S.-E. Wei, and Y.~Sheikh.
\newblock Realtime multi-person 2d pose estimation using part affinity fields.
\newblock In {\em CVPR}, 2017.

\bibitem{photographic}
Q.~Chen and V.~Koltun.
\newblock Photographic image synthesis with cascaded refinement networks.
\newblock In {\em To appear in Proceedings of the International Conference on
  Computer Vision (ICCV)}, 2017.

\bibitem{infogan}
X.~Chen, Y.~Duan, R.~Houthooft, J.~Schulman, I.~Sutskever, and P.~Abbeel.
\newblock Infogan: Interpretable representation learning by information
  maximizing generative adversarial nets.
\newblock {\em arXiv preprint arXiv:1606.03657}, 2016.

\bibitem{sketches}
M.~Eitz, J.~Hays, and M.~Alexa.
\newblock How do humans sketch objects?
\newblock {\em ACM Trans. Graph. (Proc. SIGGRAPH)}, 31(4):44:1--44:10, 2012.

\bibitem{gan}
I.~J. Goodfellow, J.~Pouget{-}Abadie, M.~Mirza, B.~Xu, D.~Warde{-}Farley,
  S.~Ozair, A.~C. Courville, and Y.~Bengio.
\newblock Generative adversarial nets.
\newblock In {\em In Neural Information Processing Systems (NIPS)}, pages
  2672--2680, 2014.

\bibitem{residual}
K.~He, X.~Zhang, S.~Ren, and J.~Sun.
\newblock Identity mappings in deep residual networks.
\newblock In {\em Computer Vision - {ECCV} 2016 - 14th European Conference,
  Amsterdam, The Netherlands, October 11-14, 2016, Proceedings, Part {IV}},
  pages 630--645, 2016.

\bibitem{pix2pix2016}
P.~Isola, J.-Y. Zhu, T.~Zhou, and A.~A. Efros.
\newblock Image-to-image translation with conditional adversarial networks.
\newblock {\em arxiv preprint arXiv:1611.07004}, 2016.

\bibitem{progressivegrowing}
T.~Karras, T.~Aila, S.~Laine, and J.~Lehtinen.
\newblock Progressive growing of gans for improved quality, stability, and
  variation.
\newblock {\em arXiv preprint arXiv:1710.10196}, 2017.

\bibitem{adam}
D.~P. Kingma and J.~Ba.
\newblock Adam: A method for stochastic optimization.
\newblock {\em CoRR}, abs/1412.6980, 2014.

\bibitem{semisup}
D.~P. Kingma, S.~Mohamed, D.~Jimenez~Rezende, and M.~Welling.
\newblock Semi-supervised learning with deep generative models.
\newblock In Z.~Ghahramani, M.~Welling, C.~Cortes, N.~D. Lawrence, and K.~Q.
  Weinberger, editors, {\em Advances in Neural Information Processing Systems
  27}, pages 3581--3589. Curran Associates, Inc., 2014.

\bibitem{vae}
D.~P. Kingma and M.~Welling.
\newblock Auto-encoding variational bayes.
\newblock {\em CoRR}, abs/1312.6114, 2013.

\bibitem{vaegan}
A.~B.~L. Larsen, S.~K. S{\o}nderby, and O.~Winther.
\newblock Autoencoding beyond pixels using a learned similarity metric.
\newblock {\em arXiv preprint arXiv:1512.09300}, 2015.

\bibitem{peopleInClothing}
C.~Lassner, G.~Pons-Moll, and P.~V. Gehler.
\newblock A generative model for people in clothing.
\newblock In {\em Proceedings of the IEEE International Conference on Computer
  Vision}, 2017.

\bibitem{srgan}
C.~Ledig, L.~Theis, F.~Huszar, J.~Caballero, A.~P. Aitken, A.~Tejani, J.~Totz,
  Z.~Wang, and W.~Shi.
\newblock Photo-realistic single image super resolution using generative
  adversarial network.
\newblock In {\em Proceedings of the IEEE Conference on Computer Vision and
  Pattern Recognition}, 2017.

\bibitem{mscoco}
T.~Lin, M.~Maire, S.~J. Belongie, L.~D. Bourdev, R.~B. Girshick, J.~Hays,
  P.~Perona, D.~Ramanan, P.~Doll{\'{a}}r, and C.~L. Zitnick.
\newblock Microsoft {COCO:} common objects in context.
\newblock {\em arXiv preprint arXiv:1405.0312}, 2014.

\bibitem{deepFashion1}
Z.~Liu, P.~Luo, S.~Qiu, X.~Wang, and X.~Tang.
\newblock Deepfashion: Powering robust clothes recognition and retrieval with
  rich annotations.
\newblock In {\em Proceedings of IEEE Conference on Computer Vision and Pattern
  Recognition (CVPR)}, 2016.

\bibitem{celeba}
Z.~Liu, P.~Luo, X.~Wang, and X.~Tang.
\newblock Deep learning face attributes in the wild.
\newblock In {\em Proceedings of International Conference on Computer Vision
  (ICCV)}, 2015.

\bibitem{deepFashion2}
Z.~Liu, S.~Yan, P.~Luo, X.~Wang, and X.~Tang.
\newblock Fashion landmark detection in the wild.
\newblock In {\em European Conference on Computer Vision (ECCV)}, 2016.

\bibitem{PoseGuidedGeneration}
L.~Ma, X.~Jia, Q.~Sun, B.~Schiele, T.~Tuytelaars, and L.~Van~Gool.
\newblock Pose guided person image generation.
\newblock In {\em To appear in Proceedings of the Conference on Advances in
  Neural Information Processing Systems (NIPS)}, pages 3846--3854, 2017.

\bibitem{milbichICCV17}
T.~Milbich, M.~Bautista, E.~Sutter, and B.~Ommer.
\newblock Unsupervised video understanding by reconciliation of posture
  similarities.
\newblock In {\em Proceedings of the IEEE International Conference on Computer
  Vision (ICCV)}, 2017.

\bibitem{acgan}
A.~Odena, C.~Olah, and J.~Shlens.
\newblock Conditional image synthesis with auxiliary classifier gans.
\newblock {\em arXiv preprint arXiv:1610.09585}, 2017.

\bibitem{dcgan}
A.~Radford, L.~Metz, and S.~Chintala.
\newblock Unsupervised representation learning with deep convolutional
  generative adversarial networks.
\newblock In {\em In International Conference On Learning Representations
  (ICLR)}, 2016.

\bibitem{whatandwhere}
S.~E. Reed, Z.~Akata, S.~Mohan, S.~Tenka, B.~Schiele, and H.~Lee.
\newblock Learning what and where to draw.
\newblock In D.~D. Lee, M.~Sugiyama, U.~V. Luxburg, I.~Guyon, and R.~Garnett,
  editors, {\em Advances in Neural Information Processing Systems 29}, pages
  217--225. Curran Associates, Inc., 2016.

\bibitem{msar}
S.~E. Reed, A.~van~den Oord, N.~Kalchbrenner, S.~G\'omez, Z.~Wang, D.~Belov,
  and N.~de~Freitas.
\newblock Parallel multiscale autoregressive density estimation.
\newblock In {\em Proceedings of The 34th International Conference on Machine
  Learning}, 2017.

\bibitem{unet}
O.~Ronneberger, P.~Fischer, and T.~Brox.
\newblock {\em U-Net: Convolutional Networks for Biomedical Image
  Segmentation}, pages 234--241.
\newblock Springer International Publishing, Cham, 2015.

\bibitem{varappforgan}
M.~Rosca, B.~Lakshminarayanan, D.~Warde{-}Farley, and S.~Mohamed.
\newblock Variational approaches for auto-encoding generative adversarial
  networks.
\newblock {\em CoRR}, abs/1706.04987, 2017.

\bibitem{rubio:PR:2015}
J.~C. Rubio, A.~Eigenstetter, and B.~Ommer.
\newblock Generative regularization with latent topics for discriminative
  object recognition.
\newblock {\em Pattern Recognition}, 48(12):3871--3880, 2015.

\bibitem{ILSVRC15}
O.~Russakovsky, J.~Deng, H.~Su, J.~Krause, S.~Satheesh, S.~Ma, Z.~Huang,
  A.~Karpathy, A.~Khosla, M.~Bernstein, A.~C. Berg, and L.~Fei-Fei.
\newblock {ImageNet Large Scale Visual Recognition Challenge}.
\newblock {\em International Journal of Computer Vision (IJCV)},
  115(3):211--252, 2015.

\bibitem{inception}
T.~Salimans, I.~J. Goodfellow, W.~Zaremba, V.~Cheung, A.~Radford, and X.~Chen.
\newblock Improved techniques for training gans.
\newblock In {\em NIPS}, 2016.

\bibitem{weightnorm}
T.~Salimans and D.~P. Kingma.
\newblock Weight normalization: A simple reparameterization to accelerate
  training of deep neural networks.
\newblock In D.~D. Lee, M.~Sugiyama, U.~V. Luxburg, I.~Guyon, and R.~Garnett,
  editors, {\em Advances in Neural Information Processing Systems 29}, pages
  901--909. Curran Associates, Inc., 2016.

\bibitem{upsample}
W.~Shi, J.~Caballero, F.~Huszar, J.~Totz, A.~P. Aitken, R.~Bishop, D.~Rueckert,
  and Z.~Wang.
\newblock Real-time single image and video super-resolution using an efficient
  sub-pixel convolutional neural network.
\newblock {\em 2016 IEEE Conference on Computer Vision and Pattern Recognition
  (CVPR)}, pages 1874--1883, 2016.

\bibitem{vgg}
K.~Simonyan and A.~Zisserman.
\newblock Very deep convolutional networks for large-scale image recognition.
\newblock {\em CoRR}, abs/1409.1556, 2014.

\bibitem{cgan}
K.~Sohn, H.~Lee, and X.~Yan.
\newblock Learning structured output representation using deep conditional
  generative models.
\newblock In {\em In Neural Information Processing Systems (NIPS)}, pages
  3483--3491, 2015.

\bibitem{pixelcnndecoder}
A.~van~den Oord, N.~Kalchbrenner, , L.~E.~K. Kavukcuoglu, O.~Vinyals, and
  A.~Graves.
\newblock Conditional image generation with pixelcnn decoders.
\newblock In {\em In Neural Information Processing Systems (NIPS)}, pages
  4790--4798, 2016.

\bibitem{ssim}
Z.~Wang, A.~C. Bovik, H.~R. Sheikh, and E.~P. Simoncelli.
\newblock Image quality assessment: From error visibility to structural
  similarity.
\newblock {\em Trans. Img. Proc.}, 13(4):600--612, Apr. 2004.

\bibitem{HED}
S.~Xie and Z.~Tu.
\newblock Holistically-nested edge detection.
\newblock In {\em In Proceedings of the IEEE International Conference on
  Computer Vision (ICCV)}, 2015.

\bibitem{attribute2image}
X.~Yan, J.~Yang, K.~Sohn, and H.~Lee.
\newblock Attribute2image: Conditional image generation from visual attributes.
\newblock In {\em Proceedings of the European Conference on Computer Vision},
  2016.

\bibitem{shoes}
A.~Yu and K.~Grauman.
\newblock Fine-grained visual comparisons wiht local learnings.
\newblock In {\em In Conference on Computer Vision and Pattern Recognition
  (CVPR)}, 2014.

\bibitem{stackgan}
H.~Zhang, T.~Xu, H.~Li, S.~Zhang, X.~Wang, X.~Huang, and D.~Metaxas.
\newblock Stackgan: Text photo-realistic image synthesis with stacked
  generative adversarial networks.
\newblock In {\em To appear in Proceedings of the International Conference on
  Computer Vision (ICCV)}, 2017.

\bibitem{pennaction}
W.~Zhang, M.~Zhu, and K.~G. Derpanis.
\newblock From actemes to action: A strongly-supervised representation for
  detailed action understanding.
\newblock In {\em Proceedings of the IEEE International Conference on Computer
  Vision}, pages 2248--2255, 2013.

\bibitem{personmultiview}
B.~Zhao, X.~Wu, Z.~Cheng, H.~Liu, and J.~Feng.
\newblock Multi-view image generation from a single-view.
\newblock {\em arXiv preprint arXiv:1704.04886}, 2017.

\bibitem{market1501}
L.~Zheng, L.~Shen, L.~Tian, S.~Wang, J.~Wang, and Q.~Tian.
\newblock Scalable person re-identification: A benchmark.
\newblock In {\em Computer Vision, IEEE International Conference on}, 2015.

\bibitem{zheng2016discriminatively}
Z.~Zheng, L.~Zheng, and Y.~Yang.
\newblock A discriminatively learned cnn embedding for person
  re-identification.
\newblock {\em arXiv preprint arXiv:1611.05666}, 2016.

\bibitem{igan}
J.-Y. Zhu, P.~Kr{\"a}henb{\"u}hl, E.~Shechtman, and A.~A. Efros.
\newblock Generative visual manipulation on the natural image manifold.
\newblock In {\em Proceedings of European Conference on Computer Vision
  (ECCV)}, 2016.

\bibitem{pix2pix_page}
J.-Y. Zhu and T.~Park.
\newblock {I}mage\-to\-{I}mage {T}ranslation with conditional adversarial nets.

\bibitem{cyclegan}
J.-Y. Zhu, T.~Park, P.~Isola, and A.~A. Efros.
\newblock Unpaired image-to-image translation using cycle-consistent
  adversarial networks.
\newblock {\em arXiv preprint arXiv:1703.10593}, 2017.

\bibitem{beYourOwnPrada}
S.~Zhu, S.~Fidler, R.~Urtasun, D.~Lin, and C.~C. Loy.
\newblock Be your own prada: Fashion synthesis with structural coherence.
\newblock In {\em Proceedings of the IEEE International Conference on Computer
  Vision}, 2017.

\end{thebibliography}
}
	
\FloatBarrier
\clearpage

\appendix
\setcounter{page}{1}
\onecolumn

\begin{center}
	\textbf{
	\Large Supplementary materials for Paper Submission 1449: \\
	\Large A Variational U-Net for Conditional Appearance and Shape Generation
	}
\end{center}

\section{Network structure}

The parameter $n$ of residual blocks in the network (see section~\ref{sec:experiments}) 
may vary for different datasets. For all experiments in the paper the value of $n$
was set to $7$.
Below we provide a detailed visualization the architecture of the model
that generates $128\times 128$ images and has $n=8$ residual blocks.

\begin{figure*}[h!]
	\begin{center}
		\includegraphics[height=0.7\textheight]{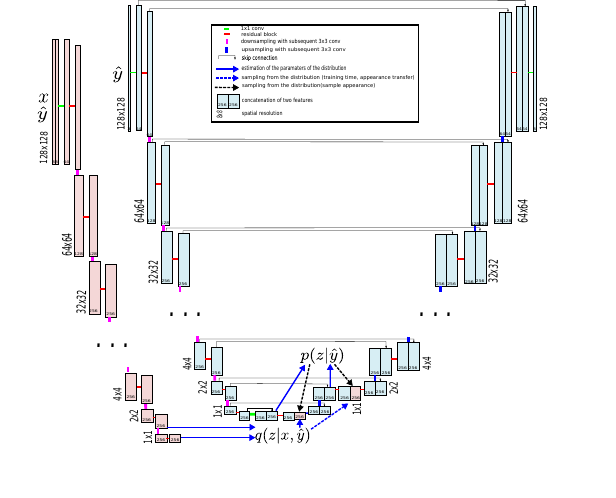}
	\end{center}
	\caption{\small{Network architecture with $8$ residual blocks for $128\times 128$ images.}}
	\label{fig:net}
\end{figure*}
\FloatBarrier

\newpage
\section{Examples of appearance sampling in different datasets}
We show more examples highlighting the ability of our model to produce
diverse samples similar to the results shown in
Fig.~\ref{fig:edges2images_samples} and \ref{fig:stickman2people_samples}.
In Fig.~\ref{fig:appearance_samples_00} we condition on edge images of shoes
and handbags and sample the appearance from the learned prior. We also run
pix2pix multiple times to compare the diversity of the produced samples. A
similar experiment is shown in Fig.~\ref{fig:appearance_samples_01}, where
we condition on human body joints instead of edge images.
	\begin{table*}[h!]
		\centering
		\begin{tabular}{c|c|ccccc|c}
			\toprule
      \small{GT} & \multicolumn{6}{c|}{\small{samples}} & \small{method} \\
			\midrule
			\multirow{ 2}{*}{\includegraphics[align=c,scale=0.1]{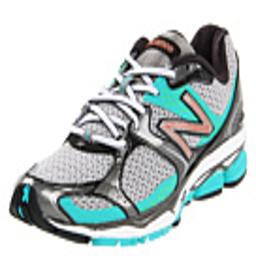}
				\includegraphics[align=c,scale=0.1]{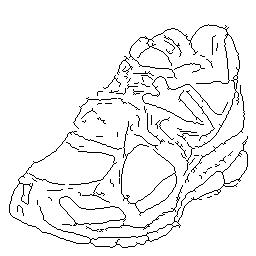}} &
			
			\includegraphics[align=c,scale=0.1]{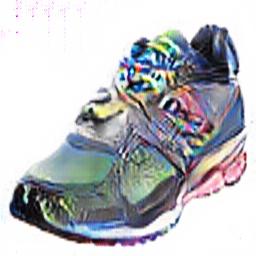} &
			\includegraphics[align=c,scale=0.1]{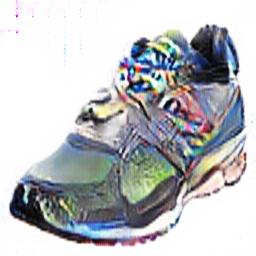} &
			\includegraphics[align=c,scale=0.1]{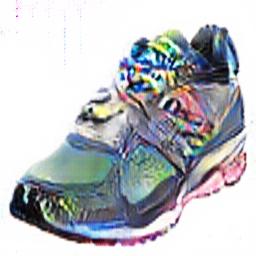} &
			\includegraphics[align=c,scale=0.1]{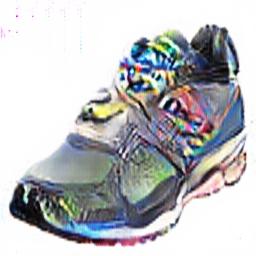} &
			\includegraphics[align=c,scale=0.1]{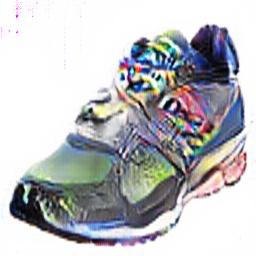} &
			\includegraphics[align=c,scale=0.1]{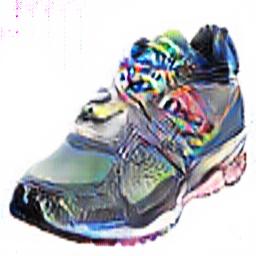} &
      pix2pix \\
			
			&
			\includegraphics[align=c,scale=0.1]{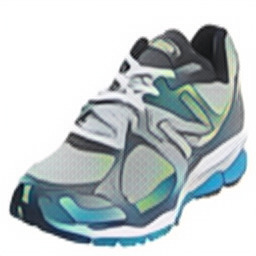} &
			\includegraphics[align=c,scale=0.1]{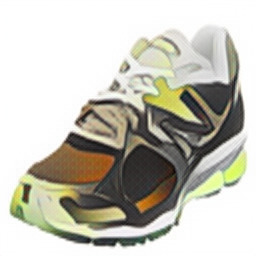} &
			\includegraphics[align=c,scale=0.1]{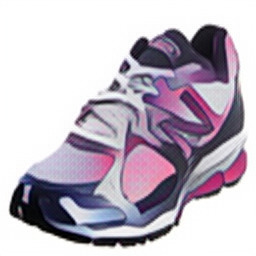} &
			\includegraphics[align=c,scale=0.1]{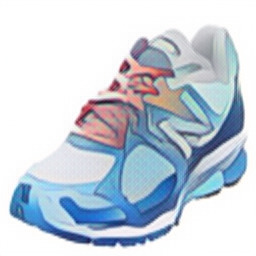} &
			\includegraphics[align=c,scale=0.1]{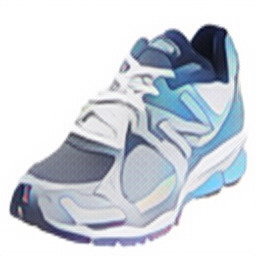} &
			\includegraphics[align=c,scale=0.1]{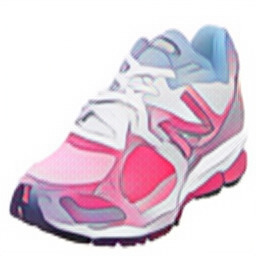} &
      our \\	

			\midrule
			\multirow{ 2}{*}{\includegraphics[align=c,scale=0.15]{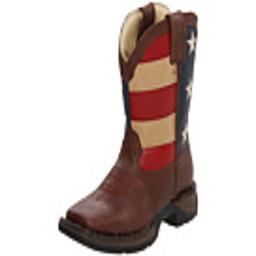}
				\includegraphics[align=c,scale=0.15]{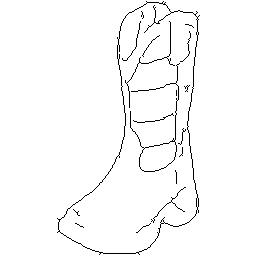}} &
			
			\includegraphics[align=c,scale=0.15]{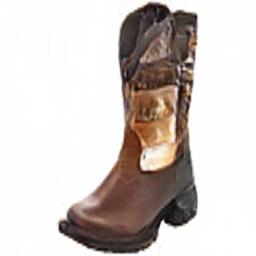} &
			\includegraphics[align=c,scale=0.15]{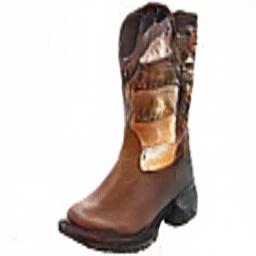} &
			\includegraphics[align=c,scale=0.15]{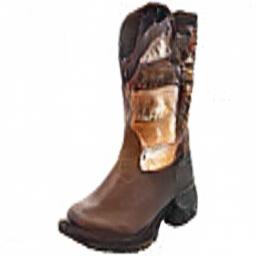} &
			\includegraphics[align=c,scale=0.15]{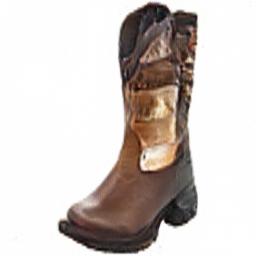} &
			\includegraphics[align=c,scale=0.15]{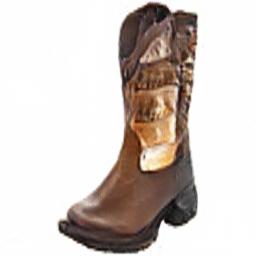} &
			\includegraphics[align=c,scale=0.15]{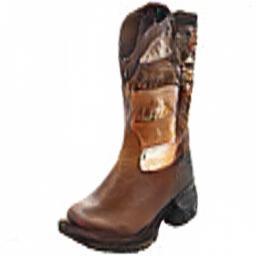} &
      pix2pix \\
			
			&
			\includegraphics[align=c,scale=0.15]{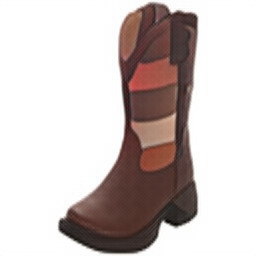} &
			\includegraphics[align=c,scale=0.15]{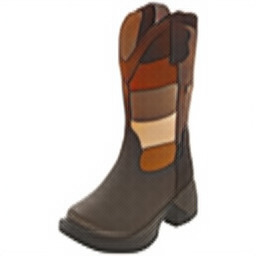} &
			\includegraphics[align=c,scale=0.15]{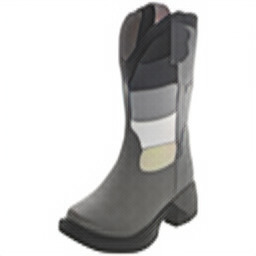} &
			\includegraphics[align=c,scale=0.15]{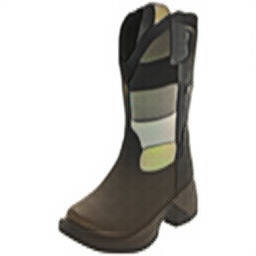} &
			\includegraphics[align=c,scale=0.15]{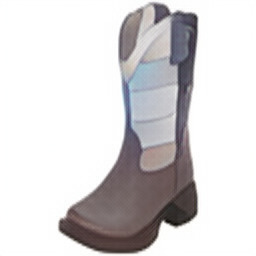} &
			\includegraphics[align=c,scale=0.15]{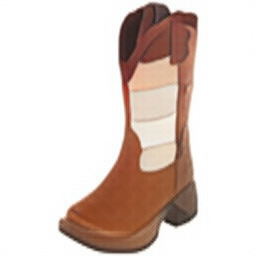} &
      our\\	
									
			\midrule
			\multirow{ 2}{*}{\includegraphics[align=c,scale=0.15]{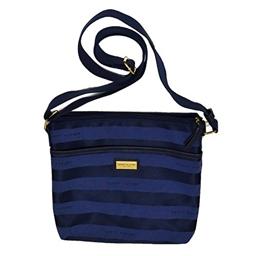}
				\includegraphics[align=c,scale=0.15]{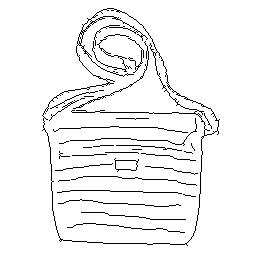}} &
			
			\includegraphics[align=c,scale=0.15]{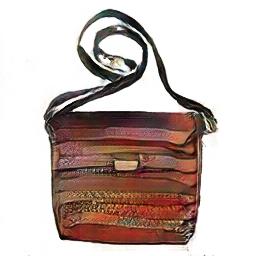} &
			\includegraphics[align=c,scale=0.15]{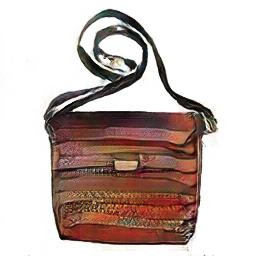} &
			\includegraphics[align=c,scale=0.15]{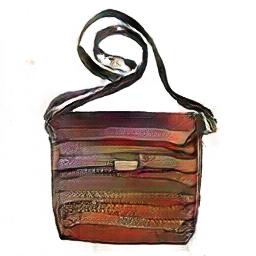} &
			\includegraphics[align=c,scale=0.15]{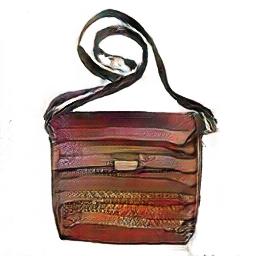} &
			\includegraphics[align=c,scale=0.15]{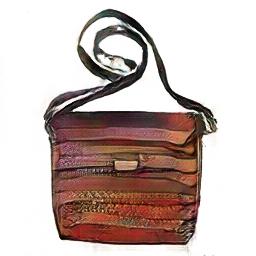} &
			\includegraphics[align=c,scale=0.15]{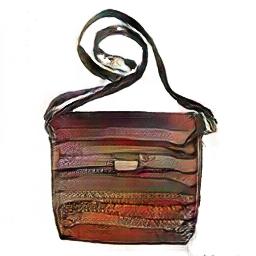} &
      pix2pix\\
			
			&
			\includegraphics[align=c,scale=0.15]{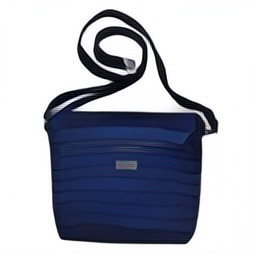} &
			\includegraphics[align=c,scale=0.15]{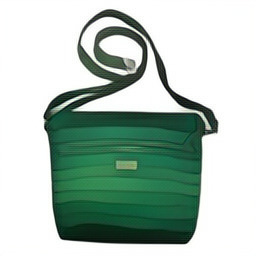} &
			\includegraphics[align=c,scale=0.15]{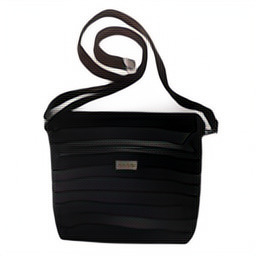} &						
			\includegraphics[align=c,scale=0.15]{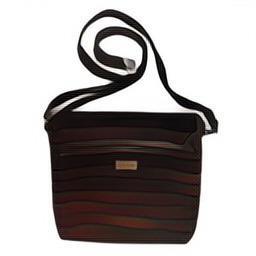} &
			\includegraphics[align=c,scale=0.15]{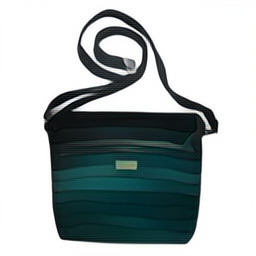} &
			\includegraphics[align=c,scale=0.15]{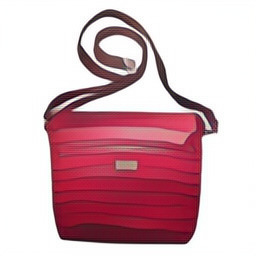} &
      our \\	
			
			\midrule
			\multirow{ 2}{*}{\includegraphics[align=c,scale=0.15]{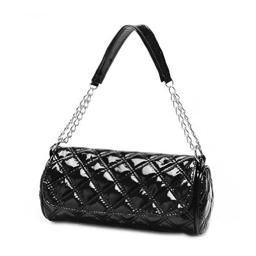}
							 \includegraphics[align=c,scale=0.15]{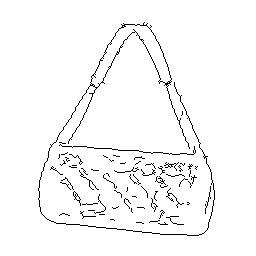}} &
			
			\includegraphics[align=c,scale=0.15]{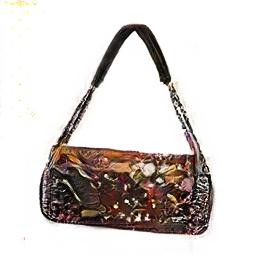} &
			\includegraphics[align=c,scale=0.15]{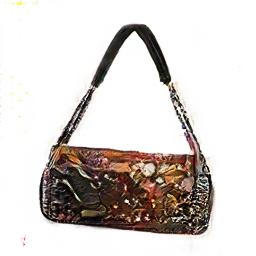} &
			\includegraphics[align=c,scale=0.15]{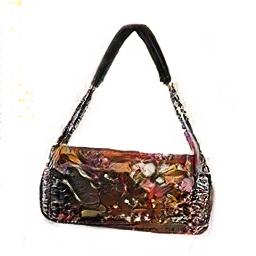} &
			\includegraphics[align=c,scale=0.15]{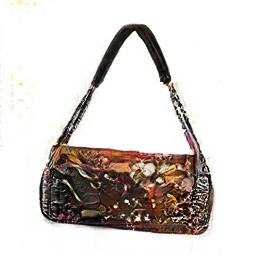} &
			\includegraphics[align=c,scale=0.15]{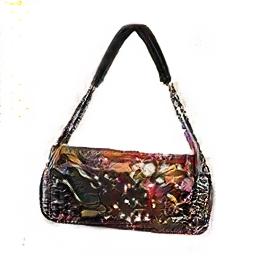} &
			\includegraphics[align=c,scale=0.15]{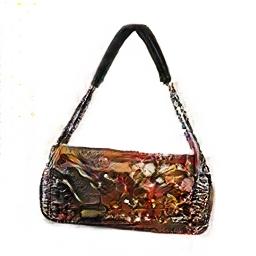} &
      pix2pix \\
			
			&
			\includegraphics[align=c,scale=0.15]{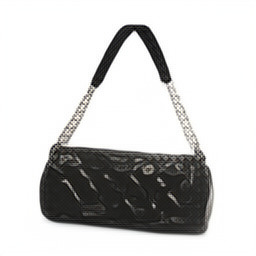} &
			\includegraphics[align=c,scale=0.15]{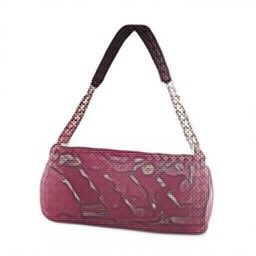} &
			\includegraphics[align=c,scale=0.15]{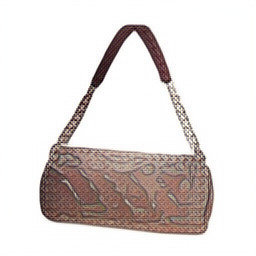} &
      \includegraphics[align=c,scale=0.15]{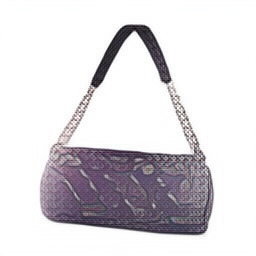} &
			\includegraphics[align=c,scale=0.15]{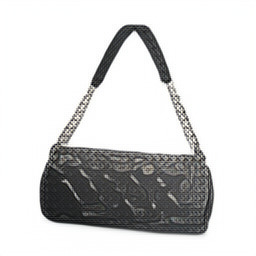} &
			\includegraphics[align=c,scale=0.15]{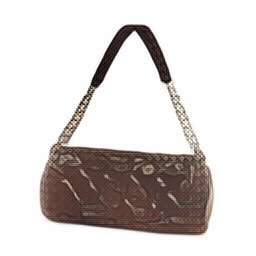} &
      our\\	
			\bottomrule	
		\end{tabular}
		\captionof{figure}{\small{Generating images based only on the edge image as
    input (GT original image and corresponding edge image are held back). We
    compare our approach with pix2pix~\cite{pix2pix2016}. On the right: each
    odd row shows images synthesized by pix2pix, each even row presents
    samples generated by our model. Here again our first image (column $2$)
    is a generation with original appearance, whereby for the $5$ following
    images we sample appearance from the learned prior distribution.
    The GT images are taken from shoes~\cite{shoes} and handbags~\cite{igan}.}} 
		\label{fig:appearance_samples_00}
	\end{table*}
	
	\begin{table*}[h!]
		\centering
		\begin{tabular}{c|c|ccccc|c}
			\toprule
      \small{GT} & \multicolumn{6}{c|}{\small{samples}} & \small{method} \\
			\midrule
			\multirow{ 2}{*}{\includegraphics[align=c,scale=0.15]{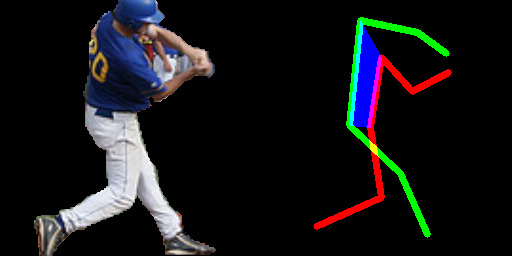}} &

			\includegraphics[align=c,scale=0.15]{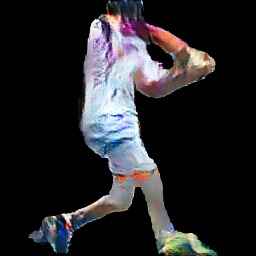} &
			\includegraphics[align=c,scale=0.15]{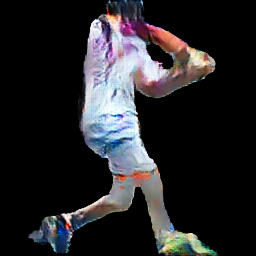} &
			\includegraphics[align=c,scale=0.15]{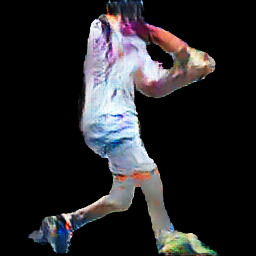} &
			\includegraphics[align=c,scale=0.15]{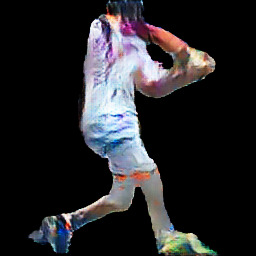} &
			\includegraphics[align=c,scale=0.15]{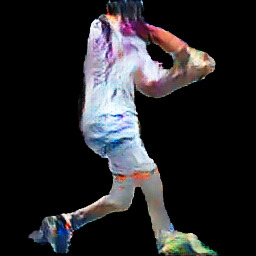} &
			\includegraphics[align=c,scale=0.15]{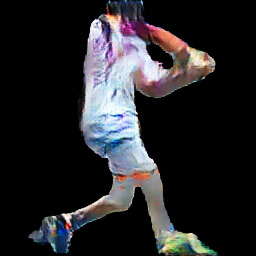}
      & pix2pix\\
			&
			\includegraphics[align=c,scale=0.15]{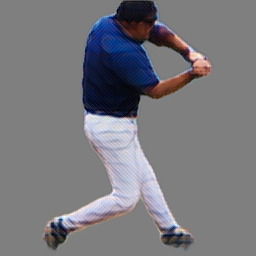} &
			\includegraphics[align=c,scale=0.15]{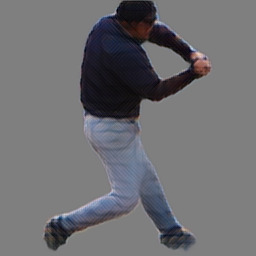} &
			\includegraphics[align=c,scale=0.15]{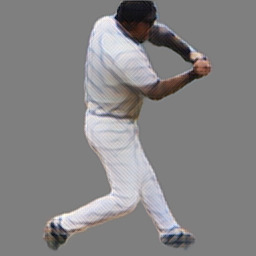} &						\includegraphics[align=c,scale=0.15]{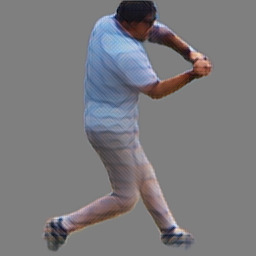} &
			\includegraphics[align=c,scale=0.15]{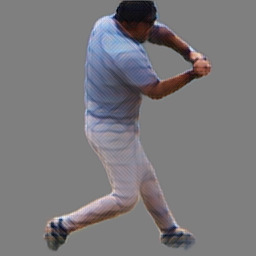} &
			\includegraphics[align=c,scale=0.15]{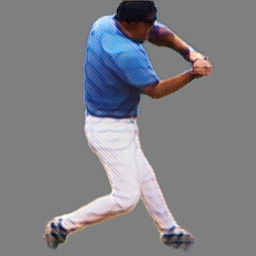}
      & our\\	
			
			\midrule
			\multirow{ 2}{*}{\includegraphics[align=c,scale=0.15]{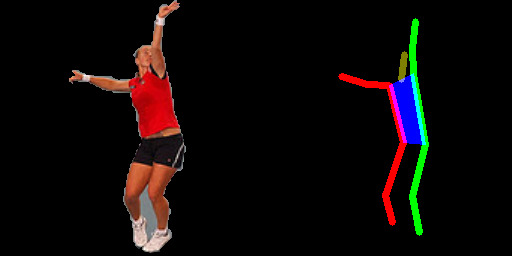}} &

			\includegraphics[align=c,scale=0.15]{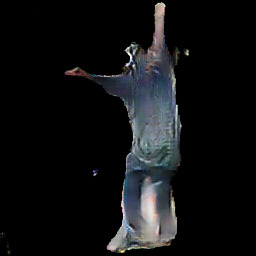} &
			\includegraphics[align=c,scale=0.15]{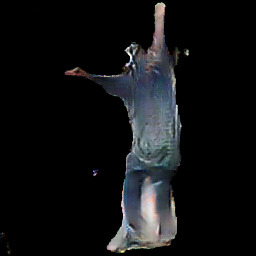} &
			\includegraphics[align=c,scale=0.15]{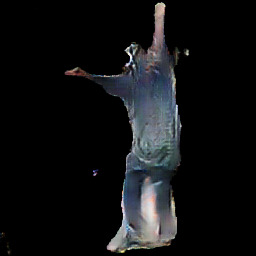} &
			\includegraphics[align=c,scale=0.15]{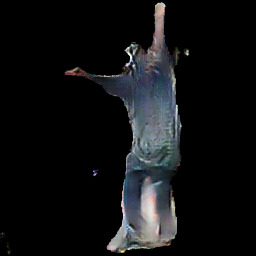} &
			\includegraphics[align=c,scale=0.15]{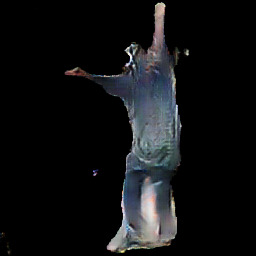} &
			\includegraphics[align=c,scale=0.15]{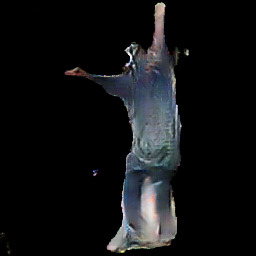}
      & pix2pix\\
			&
			\includegraphics[align=c,scale=0.15]{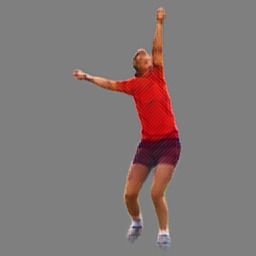} &
			\includegraphics[align=c,scale=0.15]{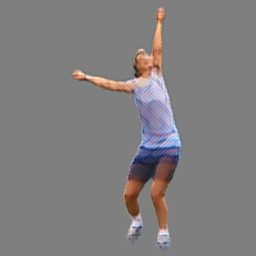} &
			\includegraphics[align=c,scale=0.15]{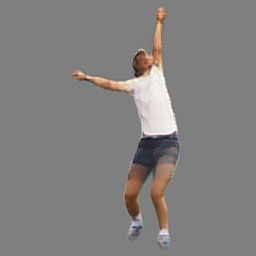} &						\includegraphics[align=c,scale=0.15]{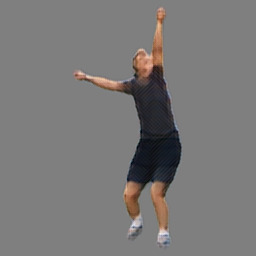} &
			\includegraphics[align=c,scale=0.15]{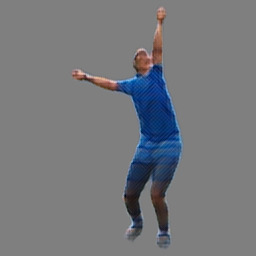} &
			\includegraphics[align=c,scale=0.15]{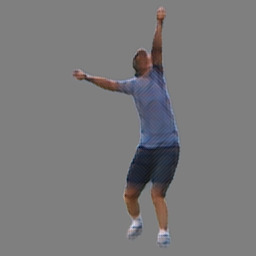}
      & our\\	
						
			\midrule
			\multirow{ 2}{*}{\includegraphics[align=c,scale=0.15]{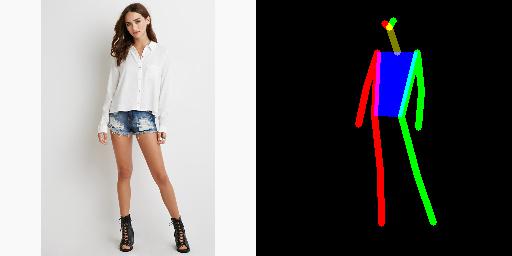}} &

			\includegraphics[align=c,scale=0.15]{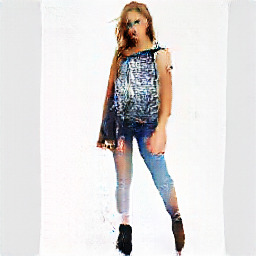} &
			\includegraphics[align=c,scale=0.15]{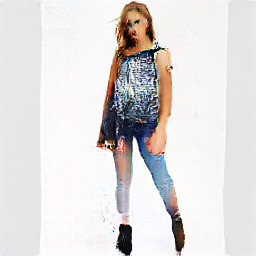} &
			\includegraphics[align=c,scale=0.15]{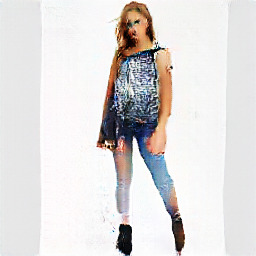} &
			\includegraphics[align=c,scale=0.15]{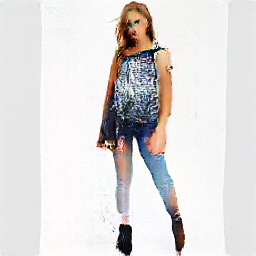} &
			\includegraphics[align=c,scale=0.15]{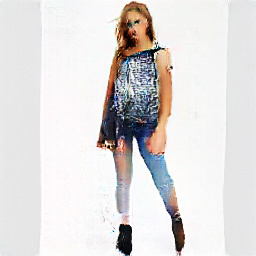} &
			\includegraphics[align=c,scale=0.15]{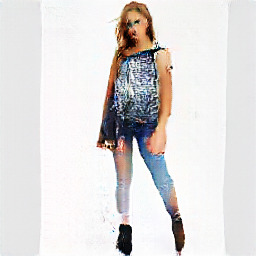}
      & pix2pix\\
			&
			\includegraphics[align=c,scale=0.15]{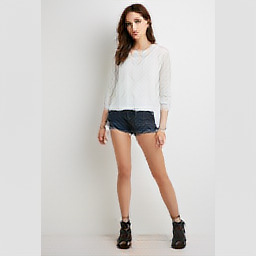} &
			\includegraphics[align=c,scale=0.15]{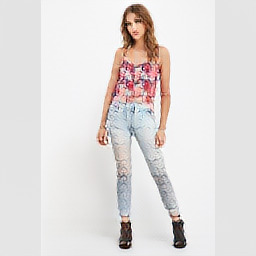} &
			\includegraphics[align=c,scale=0.15]{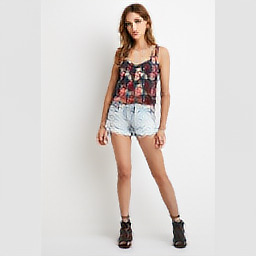} &						\includegraphics[align=c,scale=0.15]{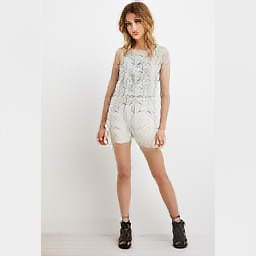} &
			\includegraphics[align=c,scale=0.15]{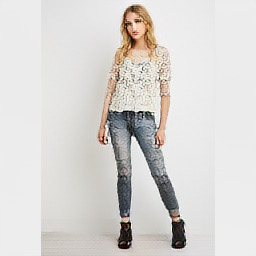} &
			\includegraphics[align=c,scale=0.15]{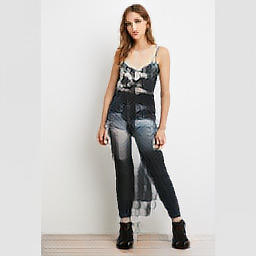}
      & our\\	
			
			\midrule
			\multirow{ 2}{*}{\includegraphics[align=c,scale=0.15]{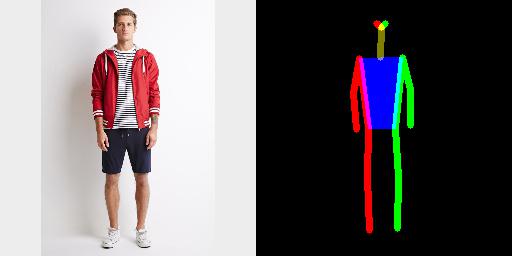}} &

			\includegraphics[align=c,scale=0.15]{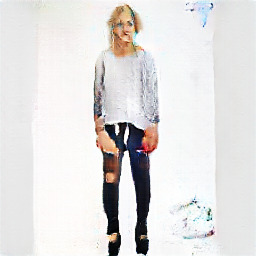} &
			\includegraphics[align=c,scale=0.15]{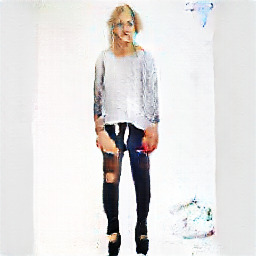} &
			\includegraphics[align=c,scale=0.15]{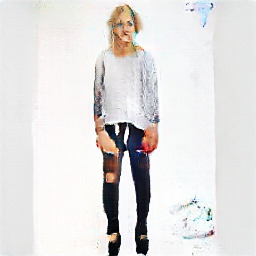} &
			\includegraphics[align=c,scale=0.15]{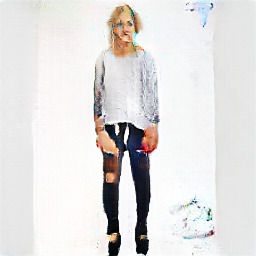} &
			\includegraphics[align=c,scale=0.15]{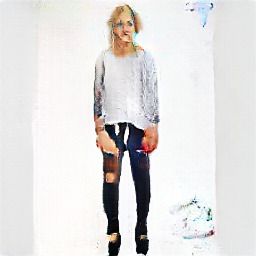} &
			\includegraphics[align=c,scale=0.15]{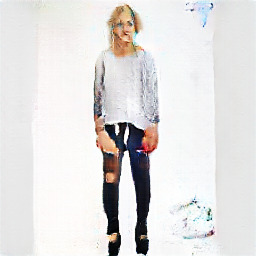}
      & pix2pix\\
			&
			\includegraphics[align=c,scale=0.15]{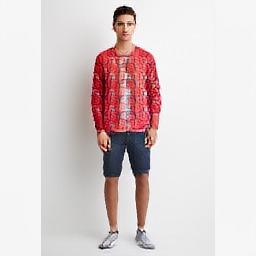} &
			\includegraphics[align=c,scale=0.15]{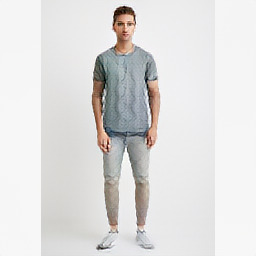} &
			\includegraphics[align=c,scale=0.15]{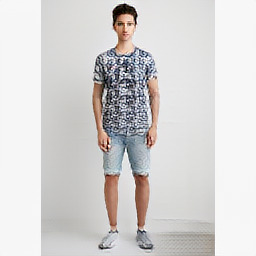} &						\includegraphics[align=c,scale=0.15]{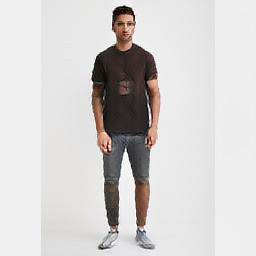} &
			\includegraphics[align=c,scale=0.15]{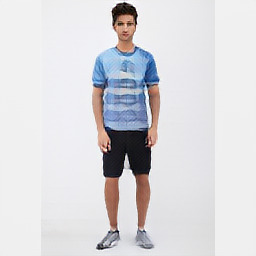} &
			\includegraphics[align=c,scale=0.15]{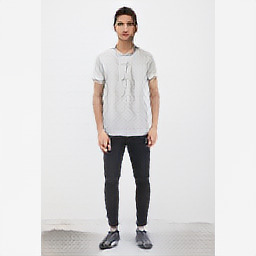}
      & our\\	
			
			\midrule
			\multirow{ 2}{*}{\includegraphics[align=c,scale=0.3]{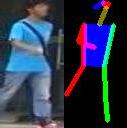}} &
			\includegraphics[align=c,scale=0.3]{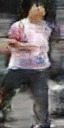} &
			\includegraphics[align=c,scale=0.3]{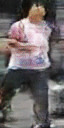} &
			\includegraphics[align=c,scale=0.3]{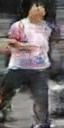} &
			\includegraphics[align=c,scale=0.3]{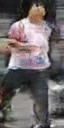} &
			\includegraphics[align=c,scale=0.3]{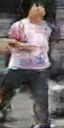} &
			\includegraphics[align=c,scale=0.3]{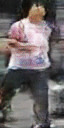}
      & pix2pix\\
			
			&
			\includegraphics[align=c,scale=0.3]{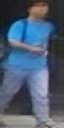} &
			\includegraphics[align=c,scale=0.3]{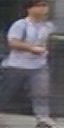} &
			\includegraphics[align=c,scale=0.3]{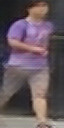} &						\includegraphics[align=c,scale=0.3]{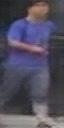} &
			\includegraphics[align=c,scale=0.3]{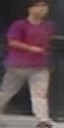} &
			\includegraphics[align=c,scale=0.3]{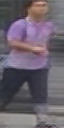}
      & our\\			
			
			\midrule
			\multirow{ 2}{*}{\includegraphics[align=c,scale=0.3]{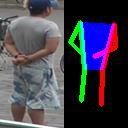}}&
			\includegraphics[align=c,scale=0.3]{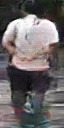} &
			\includegraphics[align=c,scale=0.3]{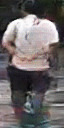} &
			\includegraphics[align=c,scale=0.3]{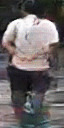} &
			\includegraphics[align=c,scale=0.3]{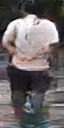} &
			\includegraphics[align=c,scale=0.3]{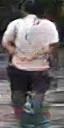} &
			\includegraphics[align=c,scale=0.3]{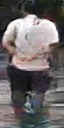}
      & pix2pix\\
			
			&
			\includegraphics[align=c,scale=0.3]{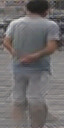} &
			\includegraphics[align=c,scale=0.3]{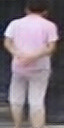} &
			\includegraphics[align=c,scale=0.3]{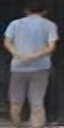} &						\includegraphics[align=c,scale=0.3]{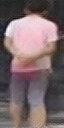} &
			\includegraphics[align=c,scale=0.3]{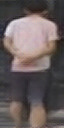} &
			\includegraphics[align=c,scale=0.3]{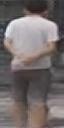}
      & our\\
			
			\bottomrule		
		\end{tabular}
		\captionof{figure}{\small{Generating images based only on the stickman as
    input (GT original image and corresponding stickman are held back). We
    compare our approach with pix2pix~\cite{pix2pix2016}. On the right: each
    odd row shows images synthesized by pix2pix, each even row presents
    samples generated by our model. Here again our first image (column $2$)
    is a generation with original appearance, whereby for the $5$ following
    images we sample appearance from the learned prior distribution.
    The GT images are taken from COCO~\cite{mscoco},
    DeepFashion~\cite{deepFashion1,deepFashion2} and
    Market-1501~\cite{market1501}.}} 
		\label{fig:appearance_samples_01}
	\end{table*}	
\FloatBarrier
	
\newpage
\section{Transfer of shape and appearance}
We show additional examples of transferring appearances to different shapes
and vice versa. We emphasize again that our approach does not require
labeled examples of images depticting the same appearance in different
shapes. This enables us to apply it on a broad range of datasets as
summarized in Table~\ref{table:overview}.

\begin{table}[h!]
	\centering
	\begin{tabular}{cccc}
		Figure & Shape Estimate & Appearance Source & Shape Target \\
		\hline
		Fig.~\ref{fig:pairs2shoes} & Edges & Handbags & Shoes \\
		Fig.~\ref{fig:pairs2handbags} & Edges & Shoes & Handbags \\
		Fig.~\ref{fig:coco} & Body Joints & COCO & COCO \\
		Fig.~\ref{fig:deepfashion} & Body Joints & DeepFashion & DeepFashion \\
		Fig.~\ref{fig:market1501} & Body Joints & Market & Market \\
		Fig.~\ref{fig:video} & Body Joints & COCO & Penn Action \\
	\end{tabular}
	\caption{Overview of transfer experiments.}
	\label{table:overview}
\end{table}

\begin{figure*}[h!]
	\begin{center}
		\includegraphics[width=0.8\textwidth]{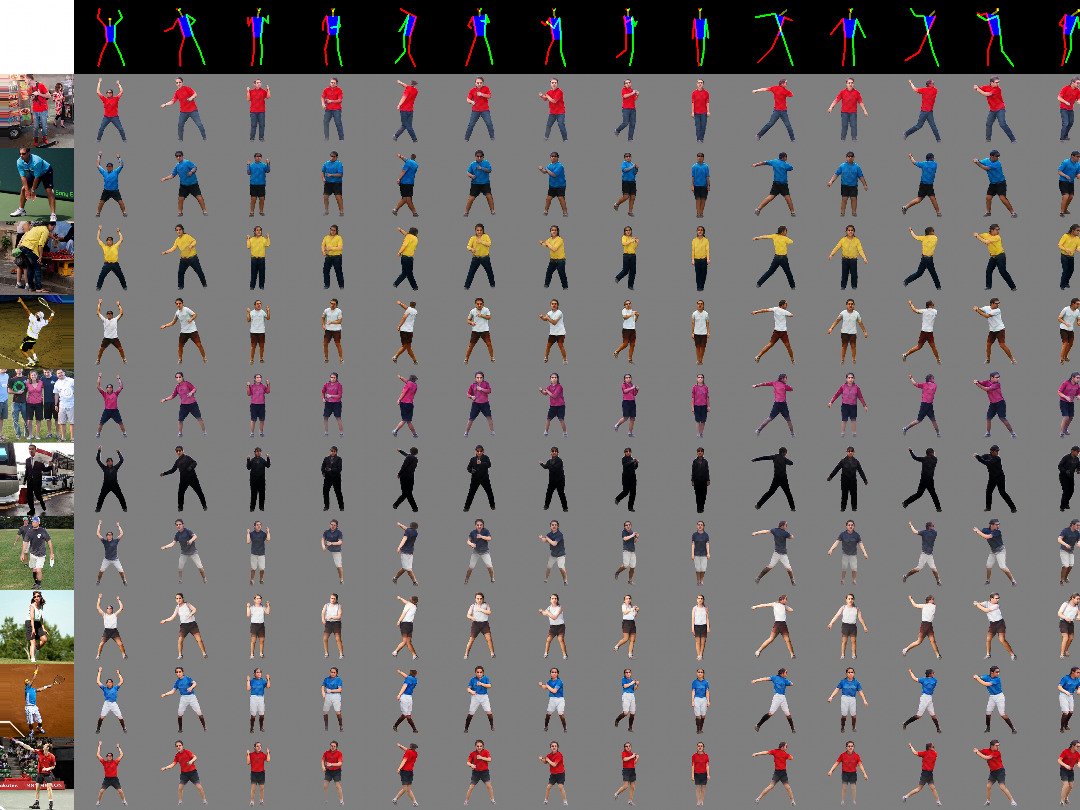}
	\end{center}
  \caption{\small{Examples of shape and appearance transfer in video.
  Appearance is inferred from COCO and target shape is estimated from Penn
  Action sequences. An animated version can be found at
  \href{https://compvis.github.io/vunet}{https://compvis.github.io/vunet}.
  Note, that we generate the video independently frame by frame without any
  temporal smoothing etc.}}
	\label{fig:video}
\end{figure*}		

	

\newpage	
\begin{figure*}[h!]
	\begin{center}
		\includegraphics[width=0.9\textwidth]{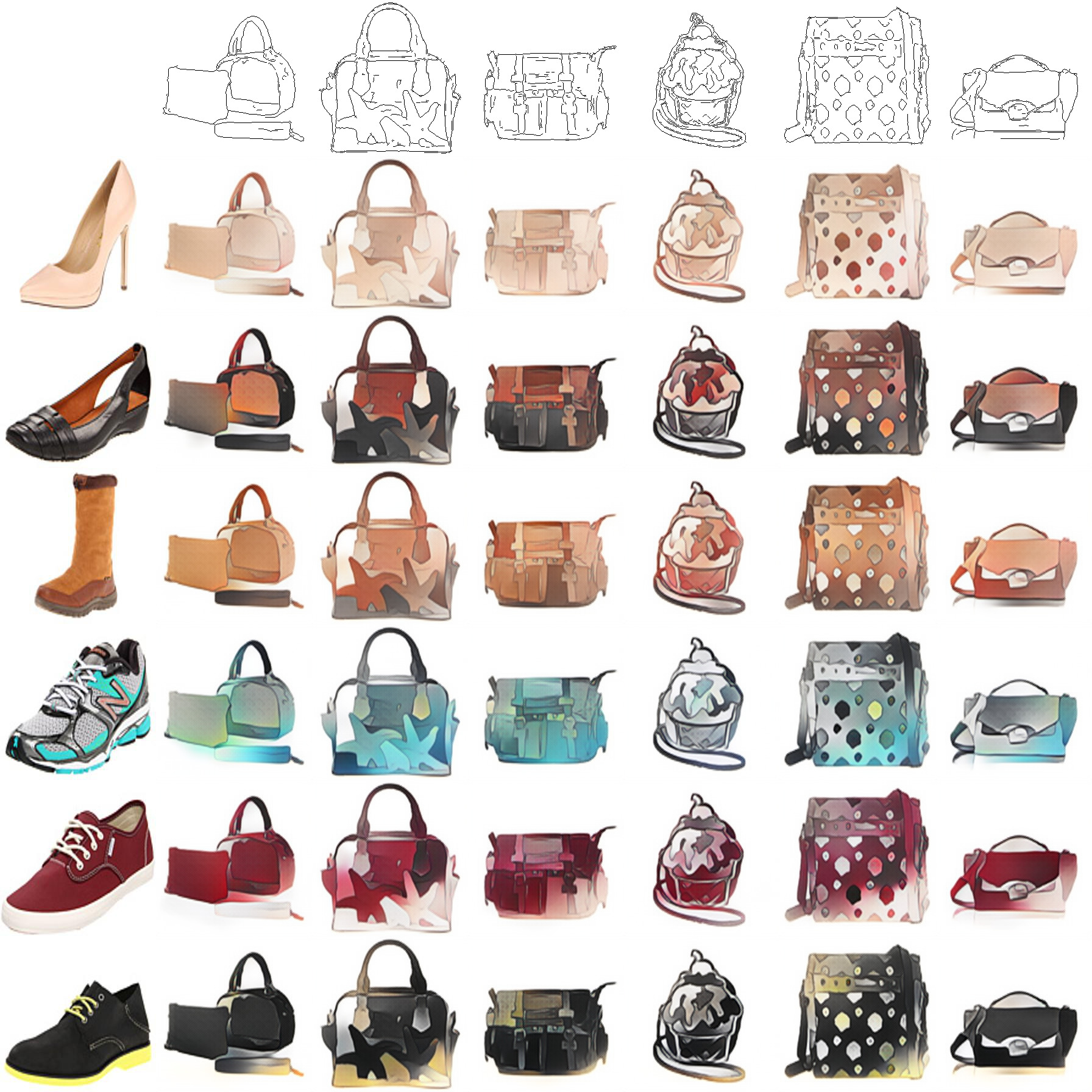}
	\end{center}
	\caption{\small{Examples of shape and appearance transfer
  between two datasets: appearance is taken from the shoes and is used to
  generate matching handbags based on their desired shape. \textit{On the
  left}: original images from the shoe dataset. \textit{On the top}: edge
  images of the desired handbags. \textit{Single row}: transfer of fixed
  appearance to different shapes. \textit{Single column}: transfer of fixed shape to different appearances.}}
	\label{fig:pairs2shoes}
\end{figure*}	
\FloatBarrier

\newpage	
\begin{figure*}[h!]
	\begin{center}
		\includegraphics[width=0.9\textwidth]{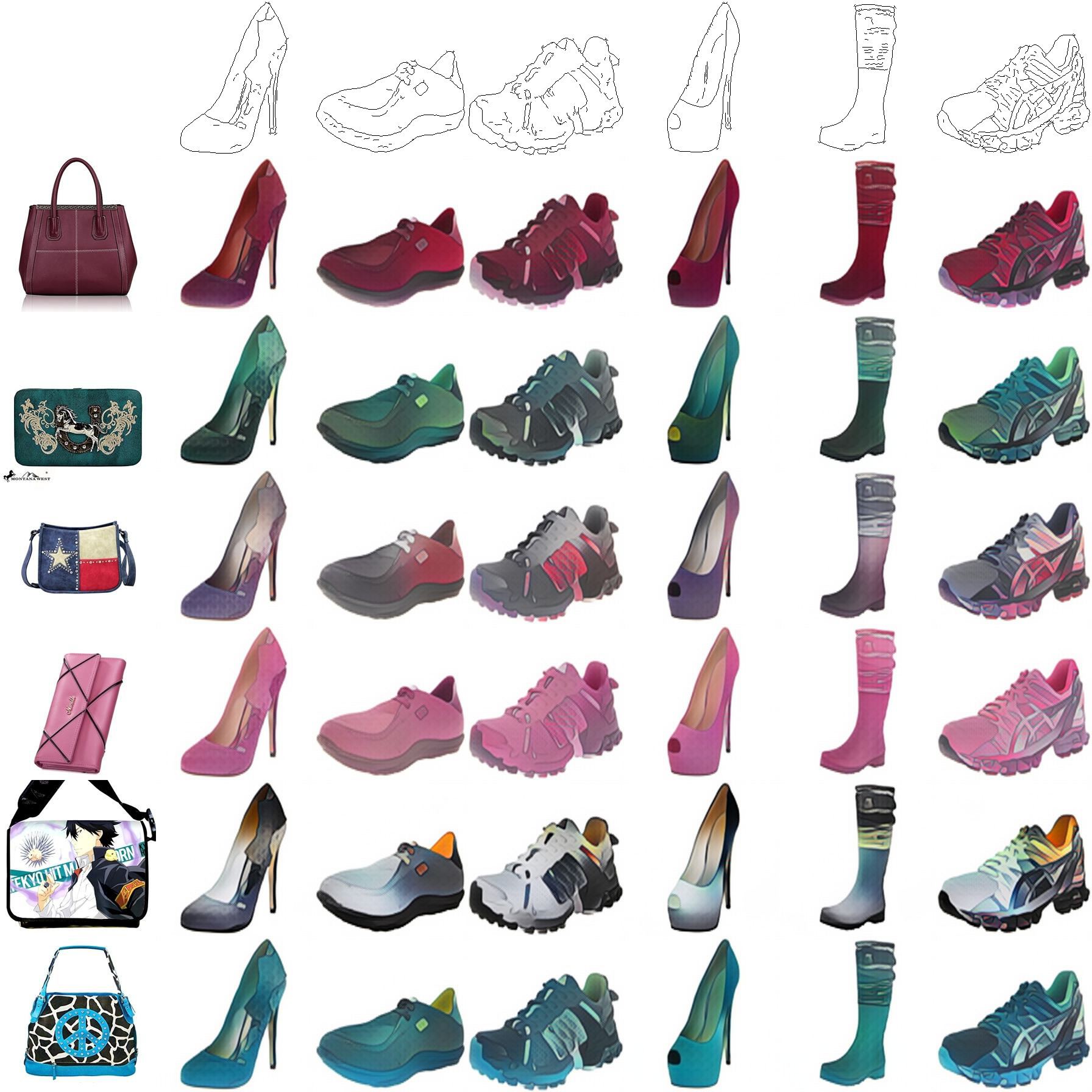}
	\end{center}
	\caption{\small{Examples of shape and appearance transfer
  between two datasets: appearance is taken from the handbags and is used to
  generate matching shoes based on their desired shape. \textit{On the
  left}: original images from the handbags dataset. \textit{On the top}:
  edge images of the desired shoes. \textit{Single row}: transfer of fixed
  appearance to different shapes. \textit{Single column}: transfer of fixed shape to different appearances.}}
	\label{fig:pairs2handbags}
\end{figure*}
\FloatBarrier

\newpage	
\begin{figure*}[h!]
	\begin{center}
		\includegraphics[width=0.9\textwidth]{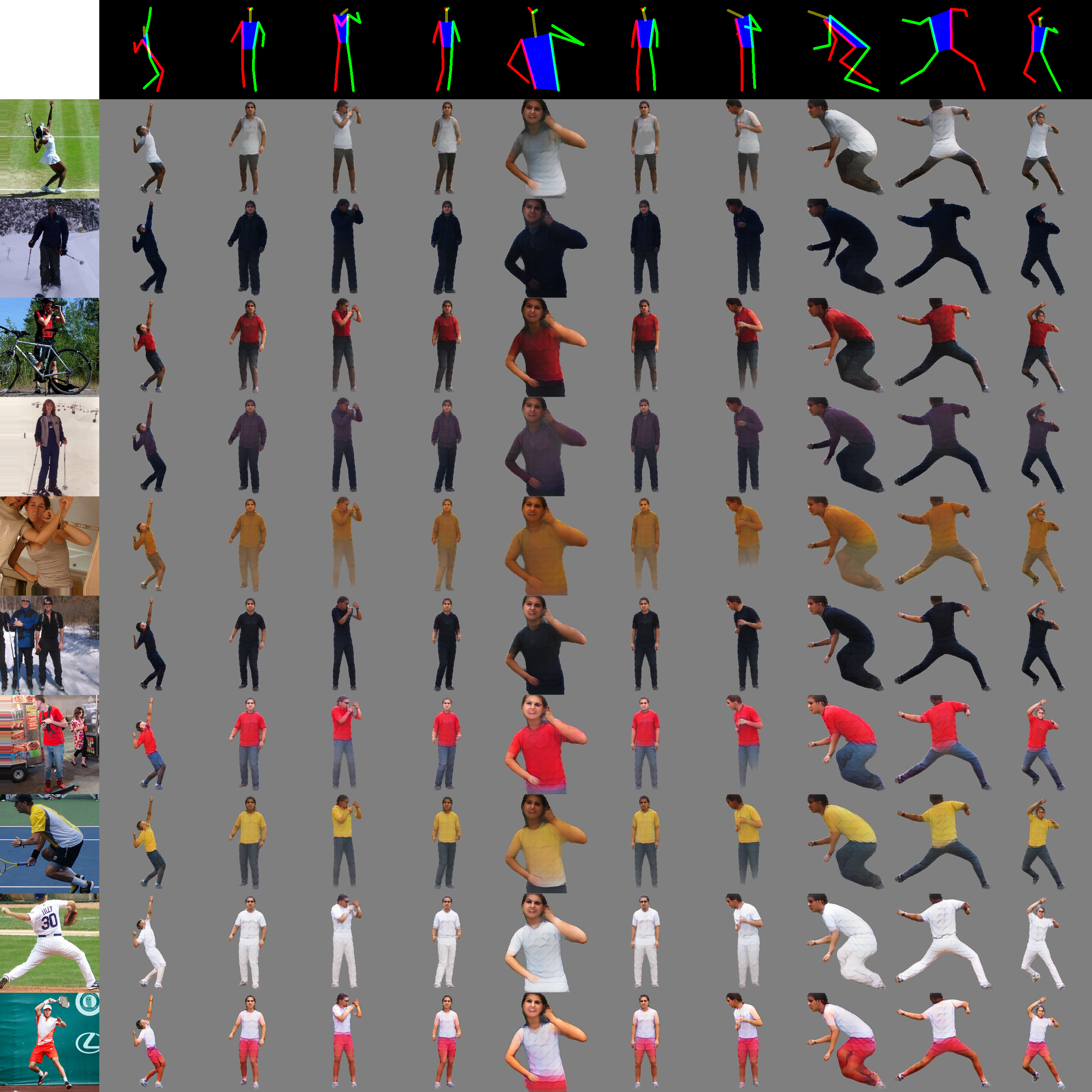}
	\end{center}
	\caption{\small{Examples of shape and appearance transfer
  on COCO dataset. \textit{On the left}: original images from the test
  split. \textit{On the top}: corresponding stickmen. \textit{Single
  row}: transfer of fixed appearance to different shapes. \textit{Single
  column}: transfer of fixed shape to different appearances.}}
	\label{fig:coco}
\end{figure*}
\FloatBarrier

\newpage	
\begin{figure*}[h!]
	\begin{center}
		\includegraphics[width=0.9\textwidth]{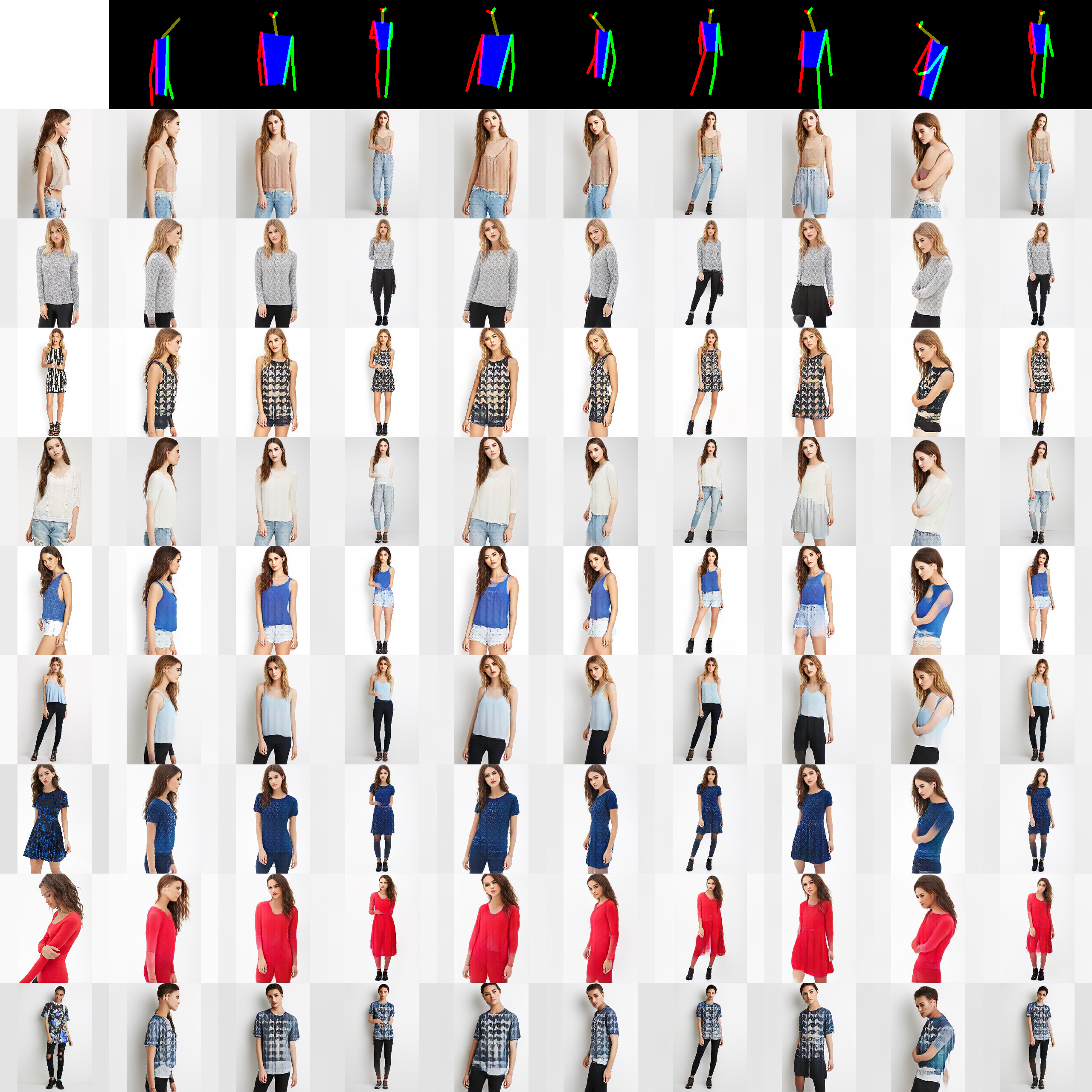}
	\end{center}
	\caption{\small{Examples of shape and appearance transfer on DeepFashion
  dataset. \textit{On the left}: original images from the test
  split. \textit{On the top}: corresponding stickmen. \textit{Single
  row}: transfer of fixed appearance to different shapes. \textit{Single
  column}: transfer of fixed shape to different appearances.}}
	\label{fig:deepfashion}
\end{figure*}	
\FloatBarrier

\newpage	
\begin{figure*}[h!]
	\begin{center}
		\includegraphics[height=0.8\textheight]{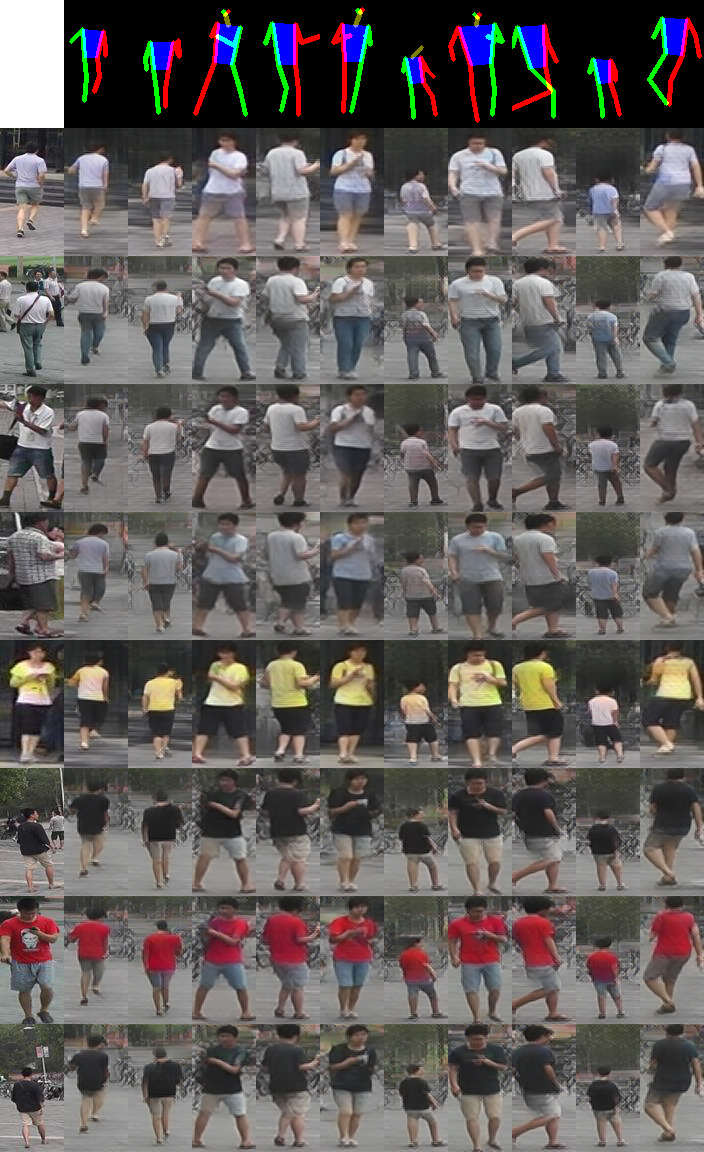}
	\end{center}
	\caption{\small{Examples of shape and appearance transfer
  on Market-1501. \textit{On the left}: original images from the test split.
  \textit{On the top}: corresponding stickmen. \textit{Single row}:
  transfer of fixed appearance to different shapes. \textit{Single column}:
  transfer of fixed shape to different appearances.}}
	\label{fig:market1501}
\end{figure*}		
\FloatBarrier

\newpage
\section{Quantitative results for the ablation study}
We have included quantitative results for the ablation study (see section~\ref{seq:Ablation})
in Table~\ref{table:is}. The positive effect of the KL-regularization
cannot be quantified by the Inception Score and thus we
presented the qualitative results in Fig.~\ref{fig:klnokl}.

\begin{table}[h!]
	\begin{center}
		\begin{tabular}{|l|c|c|c|c|}
			\hline
			method & \multicolumn{2}{c|}{Reconstruction}  & \multicolumn{2}{c|}{Transfer}\\
			\hline
			& \multicolumn{2}{c|}{IS}  & \multicolumn{2}{c|}{IS}\\
			& mean  & std & mean  & std  \\
			\hline
			our (no appearance) & 2.211 & 0.080 & 2.211 & 0.080 \\
			our (no kl) & 3.168 & 0.296 & 3.594 & 0.199 \\
			our (proposed) & 3.087 & 0.239 & 3.504 & 0.192 \\
			\hline
		\end{tabular}
	\end{center}
	\caption{\small{Inception scores (IS) for ablation study. The
			positive effect of the KL-regularization as seen in
			Fig.~\ref{fig:klnokl} cannot be quantified by the IS.}}
	\label{table:is}
\end{table}
\FloatBarrier
	
\section{Limitations}
The quality of the generated images depends highly on the dataset used for
training. Our method relies on appearance commonalities across the dataset
that can be used to learn efficient, pose-invariant encodings. If the
dataset provides sufficient support for appearance details, they are
faithfully preserved by our model (e.g. hats in DeepFashion, see Fig.~8,
third row).

The COCO dataset shows large variance in both visual qualities (e.g.
lighting conditions, resolutions, clutter and occlusion) as well as in
appearance. This leads to little overlap of appearance details in
different poses and the model focuses on aspects of appearance that can be
reused for a large variety of poses in the dataset.	

We show some failure cases of our approach in Fig.~\ref{fig:fails}.
The first row of Fig.~\ref{fig:fails} shows an example
of rare data: children are underrepresented in COCO~\cite{mscoco}. A similar
problem occurs in Market-1501~\cite{market1501} where most of the images
represent a tight crop around a person and only some contain people from
afar. This is shown in the second row which also contains an incorrect
estimate for the left leg.
Sometimes, estimated pose correlates with some other attribute of a
dataset (e.g., gender as in DeepFashion~\cite{deepFashion1,deepFashion2},
where male and female models use very characteristic yet distinct set of
poses). In this case our model morphs this attribute with the target
appearance, e.g. generates a woman with definitely male body proportions
(see row $3$ in Fig.~\ref{fig:fails}).
Under heavy viewpoint changes, appearance can be entirely unrelated, e.g.
front view showing a white t-shirt which is totally covered from the rear
view (see fourth row of Fig.~\ref{fig:fails}). The algorithm however assumes
that the appearance in both views is related.
As the example in the last row of Fig.~\ref{fig:fails} shows, our model is
confused if occluded body parts are annotated since this is not the case for
most training samples.

\begin{table}[h!]
	\centering
	\begin{tabular}{c|cc|c|c}
		               & \multicolumn{2}{c|}{\small{target shape}} & \small{target} & \\ 
		\small{reason} & \small{original image} & \small{shape estimate} &  \small{appearance} & \vtop{\hbox{\strut \small{Ours}}} \\
		\midrule
		\small{rare data} & 
		\includegraphics[align=c,width=0.15\textwidth,height=0.15\textwidth]{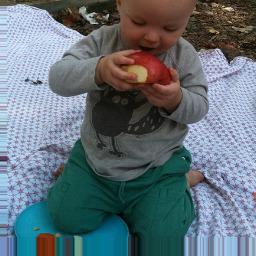} &
		\includegraphics[align=c,width=0.15\textwidth,height=0.15\textwidth]{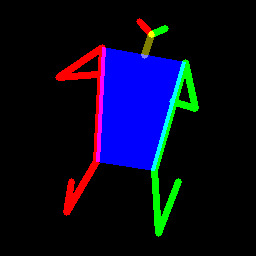} &
		\includegraphics[align=c,width=0.15\textwidth,height=0.15\textwidth]{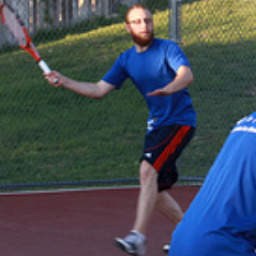} &
		\includegraphics[align=c,width=0.15\textwidth,height=0.15\textwidth]{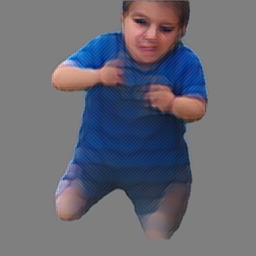} \\
		\midrule
		\vtop{\hbox{\strut \small{scale/}}\hbox{\strut \small{pose estimation error}}} & 
		\includegraphics[align=c,width=0.15\textwidth,height=0.15\textwidth]{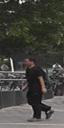} &
		\includegraphics[align=c,width=0.15\textwidth,height=0.15\textwidth]{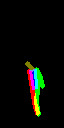} &
		\includegraphics[align=c,width=0.15\textwidth,height=0.15\textwidth]{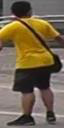} &
		\includegraphics[align=c,width=0.15\textwidth,height=0.15\textwidth]{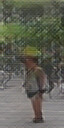} \\
		\midrule
		\vtop{\hbox{\strut \small{discriminative}}\hbox{\strut \small{pose}}} & 
		\includegraphics[align=c,width=0.15\textwidth,height=0.15\textwidth]{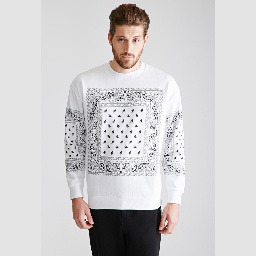} &
		\includegraphics[align=c,width=0.15\textwidth,height=0.15\textwidth]{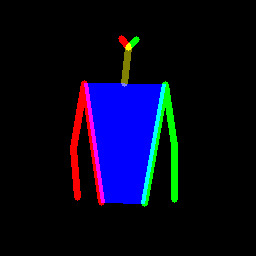} &
		\includegraphics[align=c,width=0.15\textwidth,height=0.15\textwidth]{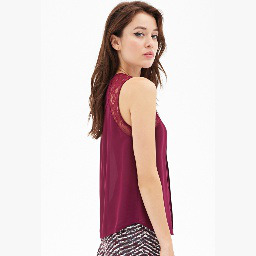} &
		\includegraphics[align=c,width=0.15\textwidth,height=0.15\textwidth]{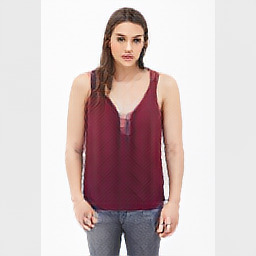} \\
		\midrule
		\vtop{\hbox{\strut \small{frontal/}}\hbox{\strut \small{backward view}}} & 
		\includegraphics[align=c,width=0.15\textwidth,height=0.15\textwidth]{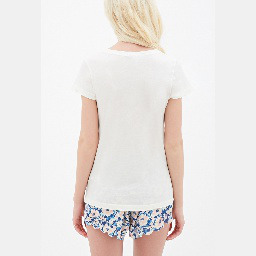} &
		\includegraphics[align=c,width=0.15\textwidth,height=0.15\textwidth]{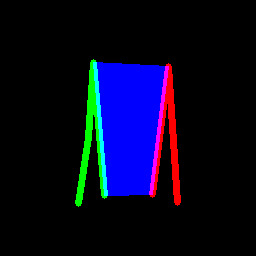} &
		\includegraphics[align=c,width=0.15\textwidth,height=0.15\textwidth]{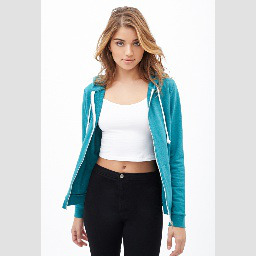} &
		\includegraphics[align=c,width=0.15\textwidth,height=0.15\textwidth]{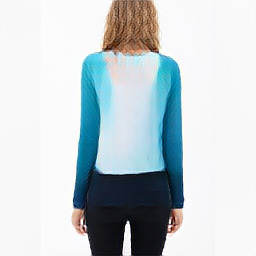} \\
		\midrule
		\vtop{\hbox{\strut \small{labeled shape}}\hbox{\strut \small{despite occlusion}}} & 
		\includegraphics[align=c,width=0.15\textwidth,height=0.15\textwidth]{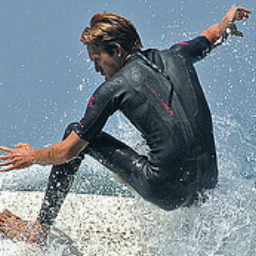} &
		\includegraphics[align=c,width=0.15\textwidth,height=0.15\textwidth]{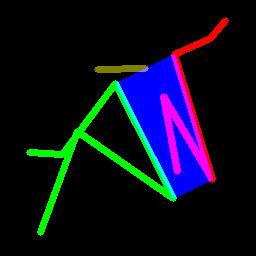} &
		\includegraphics[align=c,width=0.15\textwidth,height=0.15\textwidth]{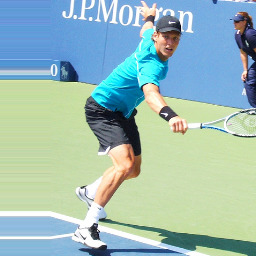} &
		\includegraphics[align=c,width=0.15\textwidth,height=0.15\textwidth]{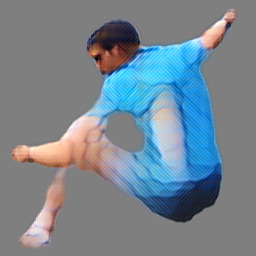} \\
	\end{tabular}
	\captionof{figure}{\small{Examples of failure cases. As most of the errors
  are dataset specific we show a collection of cases over different
  datasets.}} 
	\label{fig:fails}
\end{table}

\FloatBarrier

\end{document}